\documentclass[Afour,sageh,times]{sagej}

\usepackage{moreverb,url}

\usepackage[colorlinks,bookmarksopen,bookmarksnumbered,citecolor=red,urlcolor=red]{hyperref}
\usepackage[normalem]{ulem}

\definecolor{old_color}{RGB}{150, 150, 150}

\newcommand\BibTeX{{\rmfamily B\kern-.05em \textsc{i\kern-.025em b}\kern-.08em
T\kern-.1667em\lower.7ex\hbox{E}\kern-.125emX}}

\usepackage{graphicx, subcaption}
\usepackage{amsmath, bm} 
\usepackage{amssymb}  
\usepackage{algorithmic, xfrac}
\usepackage[linesnumbered]{algorithm2e}
\usepackage{multicol}
\usepackage{multirow}

\DeclareMathOperator*{\argmin}{arg\,min}

\newcommand{\trsp}{{\scriptscriptstyle\top}}

\begin{document}

\runninghead{Jankowski et al.}

\title{Robust Pushing: Exploiting Quasi-static Belief Dynamics and Contact-informed Optimization}

\author{Julius Jankowski\affilnum{1, 2}, Lara Bruderm\"uller\affilnum{3}, Nick Hawes\affilnum{3} and Sylvain Calinon\affilnum{1, 2}}

\affiliation{\affilnum{1}Idiap Research Institute, Switzerland\\
\affilnum{2}Ecole Polytechnique F\'ed\'erale de Lausanne, Switzerland\\
\affilnum{3}Oxford Robotics Institute, University of Oxford, UK}

\corrauth{Julius Jankowski, Idiap Research Institute,
Rue Marconi 19, 1920 Martigny, Switzerland.}

\email{julius.jankowski@idiap.ch}

\begin{abstract}
    Non-prehensile manipulation such as pushing is typically subject to uncertain, non-smooth dynamics. However, modeling the uncertainty of the dynamics typically results in intractable belief dynamics, making data-efficient planning under uncertainty difficult. This article focuses on the problem of efficiently generating robust open-loop pushing plans. First, we investigate how the belief over object configurations propagates through quasi-static contact dynamics. We exploit the simplified dynamics to predict the variance of the object configuration without sampling from a perturbation distribution. In a sampling-based trajectory optimization algorithm, the gain of the variance is constrained in order to enforce robustness of the plan. Second, we propose an informed trajectory sampling mechanism for drawing robot trajectories that are likely to make contact with the object. This sampling mechanism is shown to significantly improve chances of finding robust solutions, especially when making-and-breaking contacts is required. We demonstrate that the proposed approach is able to synthesize bi-manual pushing trajectories, resulting in successful long-horizon pushing maneuvers without exteroceptive feedback such as vision or tactile feedback. We furthermore deploy the proposed approach in a model-predictive control scheme, demonstrating additional robustness against unmodeled perturbations.
\end{abstract}


\maketitle

\section{Introduction}

Enabling robots to physically interact with the world is a key challenge in robotics. In particular, the ability to move, reorient or localize objects through contact is a fundamental capability for robots to perform tasks in unstructured environments. 
Model-based planning techniques aim to synthesize robot control actions by using a given model to reason over anticipated outcomes of actions.
However, there are two core challenges of model-based planning through contacts. First, contact dynamics are inherently discontinuous, i.e. a robot control action may have no effect on the state of a target object if the robot does not make contact. This translates into a vanishing gradient of the associated manipulation objective, making traditional gradient-based optimization difficult. 

\begin{figure}
    \centering
    \includegraphics[width=\linewidth]{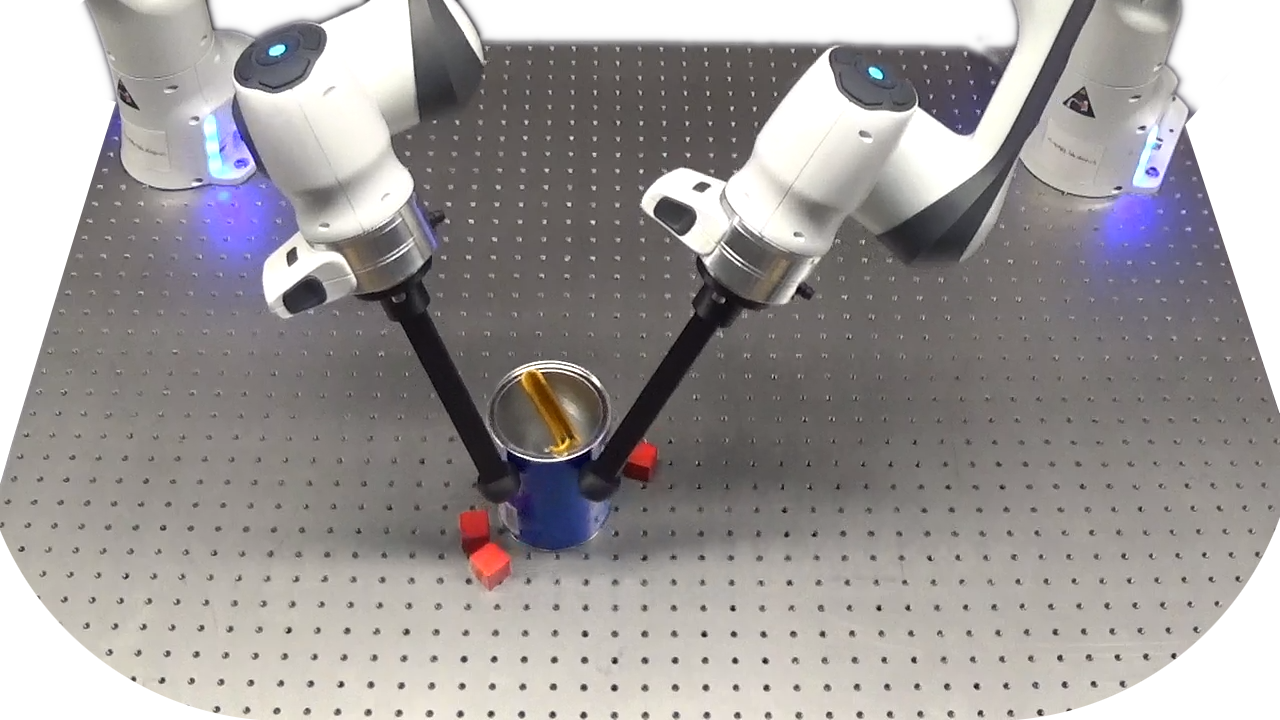}
    \caption{
        Our model-based optimization approach synthesizes bi-manual pushing trajectories by controlling the variance of the object configuration, without explicitly modeling contact modes. We show that the robustness of the pushing trajectories is sufficient to successfully push an object over long horizons. The red cubes on the table and the yellow content of the can act as additional perturbations to the contact dynamics.}
    \label{fig:shot_pushing}
\end{figure}

Second, the mechanics of non-prehensile contacts are subject to uncertainty. This is due to the fact that dynamic effects such as friction between surfaces are difficult to predict, especially when manipulating objects without knowing their physical properties, such as friction coefficients or mass distribution (\cite{Lynch1995StablePM, Dogar11, Ha20}). Thus, when generating open-loop plans by assuming an accurate model of the dynamics, those plans are likely to fail as they do not take into account the underlying uncertainty. In \cite{Rodriguez21}, this effect is described with an experiment of repeatedly picking-and-placing a queen chess piece, which fails if the queen is picked from the top. In contrast, picking the queen from the side stabilizes the repeated pick-and-place process. Uncertainty about physical parameters of objects is particularly unavoidable when robots interact with objects for the first time. This raises the question of how to let robots deal with this uncertainty autonomously. Deploying deterministic models in fast feedback loops is an implicit approach to compensating for uncertain dynamics, such as done in model-predictive control schemes. Yet they require accurate sensory measurements and achieve robustness against perturbations or modelling errors only by correcting the observed control errors.
Modeling uncertainties and reasoning over an anticipated distribution of outcomes, i.e. a belief, is a more explicit way of coping with uncertain dynamics.
Planning in belief-space enables open-loop execution of plans while still exhibiting robustness against modeled perturbations, thus not requiring sensors. Moreover, control errors are anticipated at planning time and can thus be prevented before they happen, potentially making manipulation under uncertainty more efficient. Closed-loop control in belief-space can then be achieved by continuously updating the belief based on observations and subsequent replanning, combining the benefits of reactivity with respect to unmodeled perturbations and the effectiveness of preventing control errors at planning time.

In this article, we present an approach for modeling the uncertainty of contact dynamics in order to synthesize robust manipulation behavior. Towards this end, we make the following contributions:

\noindent \textit{i)} We study how a belief over an object's configuration propagates through uncertain contact dynamics. We derive a prediction of the variance of object configurations upon contact, allowing us to anticipate the reduction of uncertainty without sampling perturbations to the contact dynamics.

\noindent \textit{ii)} We introduce a contact prior for sampling candidate robot trajectories that are likely to create contacts between the robot and the object.

\noindent \textit{iii)} Last, we propose a sampling-based trajectory optimization algorithm that constrains solutions to be robust based on the predicted variance (\textit{i)}). The informed trajectory distribution (\textit{ii)}) serves as a proposal distribution that guides the sampling-based optimization process.

In real-world experiments we demonstrate that the proposed approach is able to synthesize robust bi-manual pushing trajectories in only a few seconds of planning time, consisting of long-horizon (up to 100 seconds) open-loop pushing maneuvers that include making and breaking contacts (cf. Fig.~\ref{fig:shot_pushing}). The experimental results show that the interplay of \textit{i)} and \textit{ii)} is crucial for synthesizing robust behavior. We furthermore show that the proposed planning algorithm is fast enough to be used in a model-predictive control loop, enabling a combination of reactivity and anticipatory robustness. To the best of our knowledge, the proposed algorithm is the first model-based planning approach that is able to synthesize robust plans for contact-rich manipulation without pre-defined manipulation primitives.


\section{Related Work}

\subsection{Contact-rich Manipulation}

To plan and control physical interactions such as contacts, model-based algorithms aim to exploit these models to synthesize manipulation behavior. Optimization-based approaches have shown successful manipulation planning capabilities when the desired behavior can be found via local optimization (\cite{Hogan20, aydinoglu2022, aydinoglu2022a, cleac2021}) or when contact modes can be represented as a small set of discrete decision variables (\cite{Marcucci17, Migimatsu20, Toussaint20, Toussaint2022SequenceofConstraintsMR, Chen21}). 
In the case of local optimization, however, most methods rely on gradients which require a smoothed approximation of the contact dynamics. Moreover, they rely on good initializations, as control actions that do not result in the robot making contact with the object will yield a vanishing gradient of the associated manipulation objective.
In contrast, when reformulating the manipulation problem as finding the optimal sequence of contact modes, the difficulty lies in an combinatorial explosion when frequent switching of contact modes is necessary, i.e. making-and-breaking contacts, or if multiple contact points are involved. 

Recently, sampling-based planning and control algorithms have been explored for contact-rich manipulation tasks as a gradient-free approach to cope with discontinuous cost landscapes. This offers a framework for combining stochastic sampling and optimization, supporting the search over contact-rich manipulation actions. In \cite{Pang23}, the authors propose to use a quasi-dynamic contact model to efficiently simulate physical interactions during manipulation. Plans are then synthesized with an adaptation of the RRT algorithm. While the planning algorithm is able to generate contact-rich manipulation plans including making-and-breaking contacts, the underlying contact model is assumed to be accurate, resulting in plans that are not robust to inaccuracies in the contact model. In another stream of research on controlling contact-rich manipulation, sampling-based optimization has been explored in model-predictive control schemes to exploit parallel computing opportunities (\cite{Bhardwaj2021, howell2022predictive, jankowski2023}). In this article, we build upon our previous work on via-point-based stochastic trajectory optimization~(\emph{VP-STO}) as a tool for efficiently optimizing robot trajectories without requiring gradients of the manipulation cost with respect to the manipulation action (\cite{jankowski2023}).

\subsection{Robust Manipulation}
\begin{figure}[t]
    \centering
    \includegraphics[width=\linewidth]{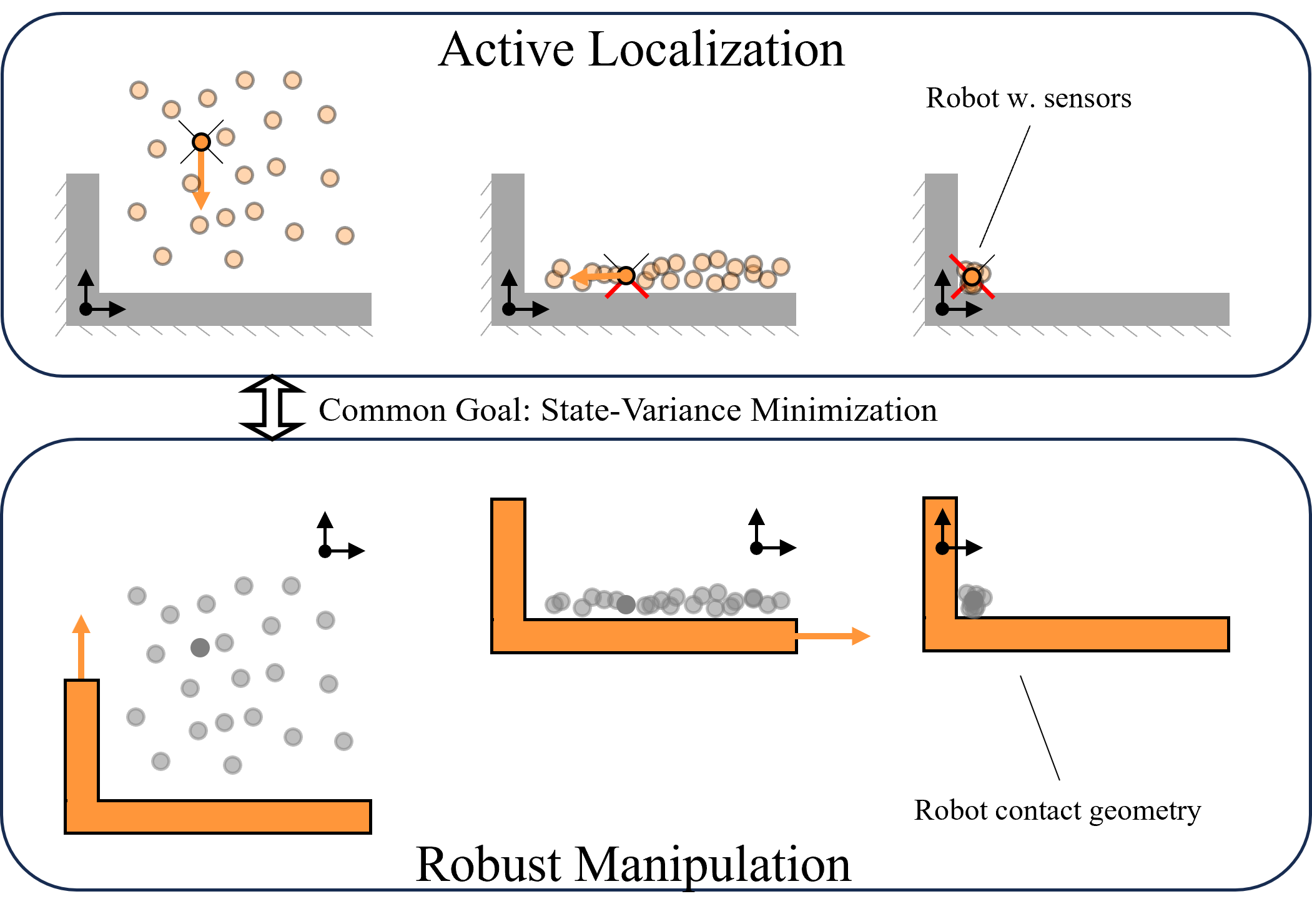}
    \caption{Analogies between active robot localization and robust manipulation. Both can be formulated as a belief space planning problem, where the objective is to decrease the uncertainty of the belief over time. While active localization approaches are based on observation models, robust manipulation exploits favorable contact dynamics to achieve the same goal.}
    \label{fig:mcl}
\end{figure}

In robotic manipulation, uncertainty is a dominant aspect that arises from the complex and hybrid nature of modeling physical interactions (\cite{Rodriguez21}). Robust manipulation aims at exploiting particular contact configurations that naturally reduce errors in manipulation tasks. In \cite{Erdmann88}, the authors exploit the geometry of a tray with physical boundaries to reorient a tool without sensory feedback. \cite{Lynch1995StablePM} exploit line contacts between a robot and polygonal objects to generate robust pushing plans. In \cite{Dogar11}, the geometry of a half-open gripper is exploited to actively funnel the probability distribution over object locations between the two fingers and the palm with a push-grasp. \cite{Ha20} model the uncertainty in an underactuated system with additive noise on the robot control actions and approximate the probability distribution over state trajectories with a normal distribution around a particular contact mode. Logic geometric programming is then used to find the most robust contact mode from a set of pre-defined candidate modes.
These strategies have in common that favorable contact geometries are used to create natural contact dynamics that effectively decrease the uncertainty of the manipulation system over time without the need for sensors. However, the above approaches rely on a pre-programmed set of robust behaviors that are tailored to the robot contact geometry.

In this article, we achieve robust manipulation as the result of optimizing over an object belief. Belief-space planning in robotics is concerned with modeling the state of the robot and its work space via probability distributions, e.g. for active localization. Sensor measurements of the robot are then used to reject or confirm possible states based on an observation model to reduce uncertainty.
Fig.~\ref{fig:mcl} illustrates the analogy of active localization and robust manipulation. Both can be formulated as a belief-space planning problem, where the common objective is to decrease the state-variance over time. While the goal is to minimize the uncertainty of the robot's own state in localization problems, robust manipulation aims to minimize the uncertainty of the object state. An even more important difference lies in the way the belief is updated over time. Instead of using observation models to rule out possible states of the robot, robust manipulation exploits natural invariances in the contact dynamics to let possible object states converge to a single state.

\subsection{Modeling Uncertainty in Contact Dynamics}

Modeling the uncertainty in contact dynamics is a key aspect for synthesizing robust robot behavior. In many reinforcement learning approaches (\cite{haarnoja2019soft, pmlr-v37-schulman15, schulman2017proximal}), domain randomization is a natural way of informing the skill learning process about all the possibilities that may be encountered when moving from simulation to reality. Domain randomization may include probability distributions over parameters of the dynamics model such as friction coefficients, object mass or object geometry (\cite{openai20, Muratore22}). By simulating a large number of combinations of policy samples and domain samples, the policy ideally converges to a behavior that is robust against the uncertainty modeled through domain randomization. We believe that domain randomization as an interface for modeling uncertainties in physical interactions is the key reason why reinforcement learning techniques show more advanced manipulation skills in the real world such as in-hand manipulation (\cite{Handa23}) compared to model-based planning and control techniques that typically assume an accurate model. Hence, we aim to bridge the gap between modeling uncertainty in contact dynamics and ad-hoc planning and control techniques that do not require data-inefficient offline training cycles by optimizing over statistical properties of a belief without sampling perturbations.


\section{Problem Formulation \& Approach} 
\label{sec:problem}

We consider an underactuated manipulation system with $n_{\mathrm{dof}}^r$ actuated degrees of freedom for the robot, and $n_{\mathrm{dof}}^o$ unactuated degrees of freedom for the object to be manipulated. We are interested in controlling the configuration of the object $\bm{q}^o \in \mathbb{R}^{n_{\mathrm{dof}}^o}$ by executing robot control commands $\bm{u} \in \mathbb{R}^{n_{\mathrm{dof}}^r}$. The initial object configuration and the contact dynamics are subject to uncertainty, such that the object configuration at time step $k$ is a random variable that is described through the \textit{belief} $b_k = p(\bm{q}^o_k)$. We formulate the planning problem as a stochastic optimal control problem over a horizon of $K$ time steps:
\begin{equation}
\label{eq:problem}
    \min_{\bm{u}_{0:K-1}} \mathrm{E}_{b_K} \left[ \left(\bm{q}^o_K - \bm{q}^o_{\mathrm{des}}\right)^\trsp \left(\bm{q}^o_K - \bm{q}^o_{\mathrm{des}}\right)\right].
\end{equation}
We are optimizing for an open-loop control trajectory $\bm{u}_{0:K-1}$ such that the expected control error at time step $K$ is minimal.
In order to evaluate the expected control error in \eqref{eq:problem} given a control trajectory, the belief over object positions is to be propagated through the stochastic contact dynamics. However, as contact dynamics are inherently non-smooth, propagating the belief through contacts in closed-form is intractable. Another way to explicitly evaluate the expected cost is to sample a large number of stochastic rollouts. Yet, this is problematic due to the fact that the evaluation of a single robot control trajectory becomes not only inefficient, but also stochastic.

A key to our approach is the separation of the stochastic optimal control problem in \eqref{eq:problem} into a mean control problem and a variance control problem (\cite{Okamoto18, Shirai23}). The objective of the stochastic optimal control problem is separated as follows:
\begin{equation}
\label{eq:problem_sep}
    \min_{\bm{u}_{0:K-1}} \left(\mathrm{E}_{b_K} \left[ \bm{q}^o_K \right] - \bm{q}^o_{\mathrm{des}}\right)^\trsp \left(\mathrm{E}_{b_K} \left[ \bm{q}^o_K \right] - \bm{q}^o_{\mathrm{des}}\right) + \mathrm{V}_{b_K} \left[ \bm{q}^o_K\right].
\end{equation}
We provide a more detailed derivation from \eqref{eq:problem} to \eqref{eq:problem_sep} in the appendix.
The first term in \eqref{eq:problem_sep} corresponds to the mean control problem, which refers to planning for the expected object configuration $\mathrm{E}_{b_K} \left[ \bm{q}^o_K \right]$, i.e. the configuration that is obtained if the mean initial configuration and the nominal contact dynamics are used. Evaluating the mean control objective does not require the propagation of the belief or the sampling of large numbers of stochastic rollouts of the state. In other words, the mean control problem resembles a deterministic optimal control problem that assumes an accurate dynamics model, while the stochasticity in the original problem is captured by the variance control problem represented by the second term in~\eqref{eq:problem_sep}. Note that the variance control problem, i.e. minimizing $\mathrm{V}_{b_K} \left[ \bm{q}^o_K\right]$, is independent of the desired object configuration $\bm{q}^o_{\mathrm{des}}$. The \textit{variance} of the object configuration is defined as: 
\begin{equation}
\label{eq:object_variance}
    \mathrm{V}_{b_k}[\bm{q}^o_k] = \mathrm{E}_{b_k}\left[\left(\bm{q}^o_k - \mathrm{E}_{b_k}[\bm{q}^o_k]\right)^\trsp \left(\bm{q}^o_k - \mathrm{E}_{b_k}[\bm{q}^o_k]\right)\right].
\end{equation}
Thus, the stochastic optimal control problem can be solved by controlling the nominal object configuration while also steering its variance. The variance is a scalar measure of the second statistical moment of the belief and can be interpreted as the uncertainty that the system has about the object's configuration.
However, note that computing the variance at time step $K$ still requires the propagation of the belief or a Monte-Carlo approximation of the stochastic dynamics. In Sec.~\ref{sec:model}, we present an approach which approximates the variance of the object configuration over time without sampling perturbations to the contact dynamics, resulting in efficient and deterministic rollouts of the nominal dynamics and the approximated variance. In Sec.~\ref{sec:bsvpsto} and Sec.~\ref{sec:multi_bsvpsto}, we exploit the approximated variance in a sampling-based trajectory optimization scheme for synthesizing robust robot trajectories.


\section{Belief Dynamics through Contacts} 
\label{sec:model}

Given a robot control action and a belief over an object's configuration, we are interested in predicting the mean and the variance of the object's configuration at the consecutive time step. For this, we first introduce a quasi-static model for the object dynamics, i.e. predicting the nominal object configuration given a robot configuration. We then model the uncertainty of the object dynamics through additive perturbations on the object configuration if the robot makes contact with the object. Last, we derive a deterministic prediction of the variance given the current belief, a control action and the statistical properties of the perturbations.

\subsection{Stochastic Quasi-Static Dynamics for Pushing}
Quasi-static and quasi-dynamic models have been used to simplify the prediction of slow physical interactions between robots and objects (\cite{Mason01, Koval16, Hogan20, Cheng21, Pang2021ACQ, Pang23}). Both classes of models assume that effects that are related to velocities and accelerations can be neglected as they do not affect the outcome of the prediction. This limits the range of applications to slow interactions such as pushing tasks, insertion tasks or in-hand manipulation. Yet, the benefits of quasi-static and quasi-dynamic models are the lower dimensionality of the system state, i.e. half the number of states compared to second-order dynamics models (e.g. \cite{mujoco}), and the lower temporal resolution required to compute stable predictions of the system state. Both aspects effectively allow for faster simulated rollouts of robot plans and thus for more efficient model-based trajectory optimization. Such models oftentimes denote the state as $\bm{q} = \left( \bm{q}^r, \bm{q}^o \right)$. It decomposes into the position of the actuated degrees of freedom of the robot $\bm{q}^r \in \mathbb{R}^{n_{\mathrm{dof}}^r}$, and the position of the unactuated degrees of freedom of the object(s) $\bm{q}^o \in \mathbb{R}^{n_{\mathrm{dof}}^o}$. The discretized dynamics are consequently given in the form of
\begin{equation}\label{eq:general_contact_dynamics}
    \begin{pmatrix} \bm{q}^r_+ \\ \bm{q}^o_+ \end{pmatrix} = \bm{f}\left(\begin{pmatrix} \bm{q}^r \\ \bm{q}^o \end{pmatrix}, \bm{u} \right).
\end{equation}
$\bm{q}^r_+$, $\bm{q}^o_+$ are the predicted robot and object configurations at the consecutive time step, respectively. In \cite{Pang23}, the input $\bm{u} \in \mathbb{R}^{n_{\mathrm{dof}}^r}$ is defined as the commanded robot configuration. The robot is assumed to be controlled by a low-level impedance controller (\cite{Hogan84}), such that the robot can be modeled as an impedance for the contact dynamics. In this article, we further simplify the quasi-dynamic contact dynamics in \cite{Pang23} by assuming infinite stiffness of the controlled robot. As a consequence, contacts with objects are assumed to not affect the robot state itself, but only the configuration of the object. This assumption is particularly realistic when pushing lightweight objects with a stiff robot impedance controller. The high-stiffness assumption induces that the robot is able to reach a desired robot position even when being in contact with an object, i.e. $\bm{q}^r_+ = \bm{u}$. The benefit of modeling the contact interactions in such a way is that the joint robot-object dynamics reduce to solely \textit{object dynamics}. This further simplifies the simulation of contacts, as the non-penetration constraint in the dynamics now only applies to the object. This turns the quasi-dynamic contact dynamics into quasi-static contact dynamics, as we remove the dependency on time. We will further exploit this decoupling of the dynamics in the subsequent derivation of transition probabilities for the object. Yet, note that this assumption breaks when contacts between the robot and the environment significantly affect the robot state, e.g. when trying to move into a solid wall. Furthermore, enclosing grasps, i.e. making contact with one object from two opposing sides, will also break the high-stiffness assumption as the two contact points will affect each other. Due to these limitations, we focus the experiments on planar pushing under uncertainty. 

\subsubsection{Nominal Object Dynamics.}

The discretized quasi-static contact dynamics of the object are
\begin{subequations}
    \begin{equation}
    \label{eq:robot_dynamics}
        \begin{pmatrix} \bm{q}^r_+ \\ \bm{q}^o_+ \end{pmatrix} = \begin{pmatrix} \bm{u} \\ \bm{q}^o + \delta \bm{q}^o \end{pmatrix};
    \end{equation}
    \begin{equation}
    \label{eq:min_object_motion}
        \delta \bm{q}^o = \argmin_{\delta \tilde{\bm{q}}^o} {\delta \tilde{\bm{q}}^o}^\trsp \bm{M}(\bm{q}^o)  \delta \tilde{\bm{q}}^o,
    \end{equation}
    \begin{equation}
    \label{eq:penetration_constraint}
       \mathrm{s.t.} \quad d(\bm{u}, \bm{q}^o + \delta \tilde{\bm{q}}^o) \geq 0.
    \end{equation}
\end{subequations}
The control inputs $\bm{u} \in \mathbb{R}^{n_{\mathrm{dof}}^r}$ to the system dynamics are defined by the commanded joint positions of the robot.
In \eqref{eq:min_object_motion}, $\bm{M}(\bm{q}^o)$ is the inertia matrix of the object with respect to its current configuration $\bm{q}^o$. Thus, the objective in \eqref{eq:min_object_motion} aims to minimize the work required to overcome the friction between the object surface and the surface of the environment, e.g. the table the object is placed on.
$d(\bm{q}^r, \bm{q}^o)$ measures the shortest signed distance between the robot and the object and thus \eqref{eq:penetration_constraint} incorporates the non-penetration constraint, i.e. the robot does not penetrate a rigid object.
As the quasi-static robot dynamics in \eqref{eq:robot_dynamics} are decoupled from the object configuration, the non-penetration constraint in \eqref{eq:penetration_constraint} does not depend on the previous robot state, but only on the control input $\bm{u}$.
This simplification allows us to represent the system dynamics with the next desired robot position as the control input and the object position as the sole state variable.
The quasi-static contact dynamics of the object in \eqref{eq:robot_dynamics}--\eqref{eq:penetration_constraint} can thus be summarized into the nominal forward dynamics of the object configuration:
\begin{equation}
\label{eq:object_dynamics}
    \bm{q}^o_+ = \bm{f}(\bm{q}^o, \bm{u}).
\end{equation}

In the following, we exploit the simplified mathematical structure of the quasi-static contact dynamics to model uncertainty and to analyze how object belief states propagate through the contact dynamics.

\subsubsection{Noisy Object Dynamics.}

We propose to model the uncertainty in the contact dynamics as additive noise that acts as a perturbation to the nominal quasi-static contact dynamics in \eqref{eq:object_dynamics}. Hence, the noisy object dynamics are given by
\begin{equation}
\label{eq:object_dynamics_noisy}
    \bm{q}^o_{+} = \bm{f}(\bm{q}^o, \bm{u}) + \eta \bm{w}.
\end{equation}
The perturbation $\bm{w}$ is assumed to be sampled from a probability distribution $p_{\bm{w}}$ such that the resulting object configuration satisfies the non-penetration constraint in \eqref{eq:penetration_constraint}. The noise coefficient $\eta$ encodes if the robot is in contact with the object and is defined as
\begin{equation}
    \eta = \begin{cases}
        0 \quad \quad \mathrm{if} \, d (\bm{u}, \bm{q}^o) > 0 \; \mathrm{(no\;contact)}, \\
        1 \hfill \mathrm{else} \; \mathrm{(contact)}.
    \end{cases}
\end{equation}
Hence, the model only considers uncertainty in the prediction of the object configuration when the robot is manipulating the object. The presented formulation of the noisy discrete-time contact dynamics in \eqref{eq:object_dynamics_noisy} captures two distinct modes: one when the robot is in contact with the object ($\eta = 1$) and another when the robot is not in contact with the object ($\eta = 0$). Note that we do not model different contact modes. The contact mode $\eta = 1$ captures arbitrary numbers of contact points with arbitrary contact geometries. The object dynamics can thus be formulated with
\begin{equation}
\label{eq:bimodal_transition}
    \bm{q}^o_{+} = \begin{cases} \bm{q}^o \quad \quad \quad \mathrm{if} \, \eta = 0 \; \mathrm{(no\;contact)}, \\
        \bm{f}(\bm{q}^o, \bm{u}) + \bm{w} \hfill \mathrm{else} \; \mathrm{(contact)}.
    \end{cases}
\end{equation}
The \textit{no-contact} mode entails that the object configuration does not change. This mode is not subject to uncertainty.



\subsection{Object Belief Dynamics}

\begin{figure}
    \centering
    \includegraphics[width=\linewidth]{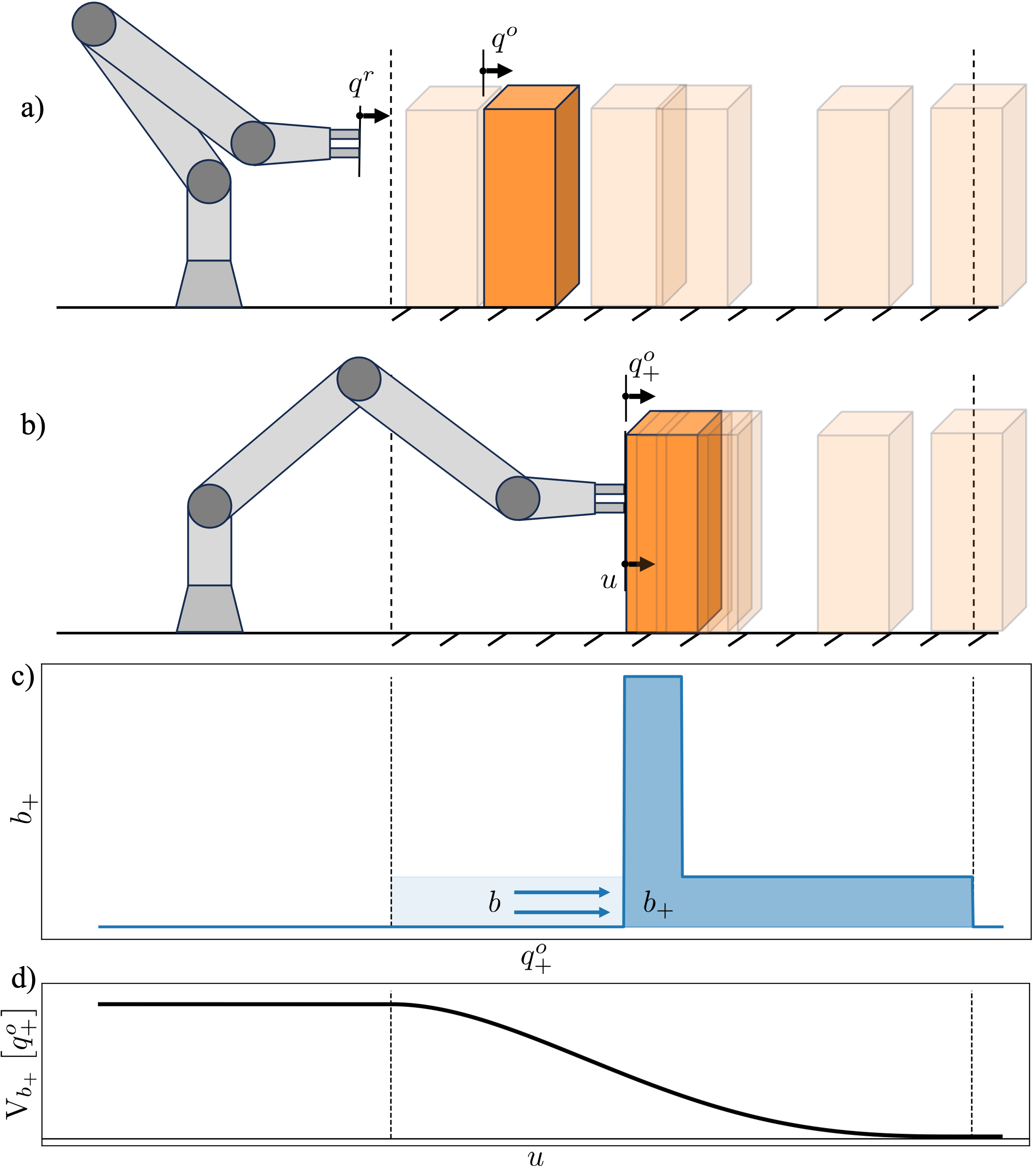}
    \caption{Belief dynamics through contact in a one-dimensional example. a) Illustration of object samples from the uniformly distributed initial belief (orange). The ground truth object is depicted in non-transparent orange. b) After executing a push from left to right, all samples that were on the left of the push were pushed by the robot. c) The probability mass that was on the left-hand side of the push (light-blue) is now concentrated in a distribution at the contact point according to the perturbation distribution (blue). The probability mass on the right-hand side of the push does not change (blue). d) The variance of the predicted object position is therefore a function of the control action, where a robust control action may decrease the variance over time.}
    \label{fig:finger_push_1D}
\end{figure}

In order to analyze how the belief about an object configuration and its associated uncertainty propagate through contact dynamics over time, we use the notation of probabilistic state transitions. Building upon the derived contact dynamics in \eqref{eq:object_dynamics_noisy}, the state of the stochastic system is the object configuration and the action is the next desired robot configuration. Accordingly, we denote the state transition probability with
\begin{equation}
\label{eq:transition}
    \bm{q}^o_{+} \sim p(\cdot | \bm{q}^o, \bm{u}),
\end{equation}
describing the probability distribution over object configurations at the next time step given the commanded robot configuration as well as the object configuration. Note that the random perturbation $\bm{w}$ that directly acts on the object dynamics in \eqref{eq:object_dynamics_noisy} is captured by the stochasticity of the transition probability in \eqref{eq:transition}. We furthermore denote the probability distribution over object configurations as \textit{belief} $b = p(\bm{q}^o)$.
Given the transition probability in \eqref{eq:transition}, a control action $\bm{u}$ and the object belief $b$, the resulting belief $b_+$ can be predicted by marginalizing over the initial object configuration:
\begin{equation}
\label{eq:belief_dynamics}
    b_{+} = p(\bm{q}^o_{+} | \bm{u}) = \int_{\mathcal{Q}^o} p(\bm{q}^o_{+} | \bm{q}^o, \bm{u}) \;  b \; \mathrm{d}\bm{q}^o.
\end{equation}
This equation represents the \textit{belief dynamics}, which can be used not only for predicting the most likely object configuration, but also the variance of the belief after executing action $\bm{u}$. However, solving the integral in \eqref{eq:belief_dynamics} is typically intractable for non-linear dynamics and non-Gaussian beliefs. Yet, in the following, we start by showing an example problem for which we can obtain a closed-form solution of \eqref{eq:belief_dynamics}, followed by the introduction of a general approximation. \newline

\noindent
\textbf{Example 1. (1D Box-Pushing).} We illustrate the effects of contacts on the belief with a one-dimensional pushing example as illustrated in Fig.~\ref{fig:finger_push_1D}. The hand (i.e. the robot) can be controlled directly while the box on the table (i.e. the object) can only be moved by making contact. The quasi-static contact dynamics in this example simplify to
\begin{equation}
    q^o_{+} = \begin{cases}
        q^o \quad \, \mathrm{if} \, q^o > u \; \mathrm{(no\;contact)}, \\
        u + w \quad \mathrm{else} \hfill \mathrm{(contact)}.
    \end{cases}
\end{equation}
These one-dimensional piece-wise linear dynamics move the object to the right-hand side of the contact. The object does not move if no contact has been made. In this example, we consider the perturbation of the contact dynamics to be uniformly distributed, i.e. $w \sim \mathcal{U}_{[0, \alpha]}$. Sub-figures a) and b) illustrate a particular instance of the object being at $q^o$, which is then moved to $q^o_{+}$ due to the robot reaching the commanded position $u$. However, in this example the initial box position is subject to uncertainty. The initial belief $b$ over box positions being given as a uniform distribution on an interval between box position $\underline{q}^o$ and $\bar{q}^o$, respectively. The interval is indicated by the vertical dashed lines in all sub-figures. The initial belief is
\begin{equation}
    b = \mathcal{U}_{[\underline{q}^o, \bar{q}^o]}(q^o).
\end{equation}
Given a control action $u \in [\underline{q}^o, \bar{q}^o]$ as a commanded robot position, we can now predict the belief after contact by solving \eqref{eq:belief_dynamics}. While solving the integral in \eqref{eq:belief_dynamics} is typically intractable, this example has a closed-form solution that is given by
\begin{equation}
\label{eq:example_prediction}
b_+ = \frac{u - \underline{q}^o}{\bar{q}^o - \underline{q}^o} \mathcal{U}_{[u, u + \alpha]}(q^o_+) + \frac{\bar{q}^o - u}{\bar{q}^o - \underline{q}^o} \mathcal{U}_{[u, \bar{q}^o]}(q^o_+).
\end{equation}
We illustrate the belief dynamics in sub-figure c) by visualizing the initial belief in light-blue and the resulting belief in dark-blue. It can be seen that the contact dynamics result in a concentration of probability mass around the contact point. This is due to the fact that all probability mass to the left of the contact has moved to the same position interval, i.e. the contact point subject to perturbation. In the extreme case of pushing all the way through the interval of the uniform distribution, i.e. $u = \bar{q}^o$, the object belief is equivalent to the perturbation distribution with the mean shifted to the contact point.
The closed-form belief dynamics in \eqref{eq:example_prediction} show that the probability distribution can be controlled by the actions a robot takes. This is underlined by sub-figure d) plotting the variance $\mathrm{V}$ of the object position after the push as a function of how far the robot pushed. Note that while in this specific example any control action $u$ that makes contact with the object reduces the variance of the belief over its position, this may not be the case in general. An unfavorable control action could generally also increase the variance of the belief. 


\subsection{Variance Prediction.}
\label{sec:variance_dynamics}
While the 1D-example above showed a closed-form solution to the belief dynamics, this is typically intractable. Consequently, predicting the variance with
\begin{equation}
\label{eq:variance_pred_intractable}
    \mathrm{V}_+ = \mathrm{V}_{b_+}\left[ \bm{q}^o \right],
\end{equation}
in closed-form is also intractable. A common approach is to use a Monte-Carlo approximation of the belief dynamics in \eqref{eq:belief_dynamics} in order to compute the variance of the approximated belief update (\cite{Kappen2015AdaptiveIS, Shirai23}). 
However, this is problematic since this induces stochasticity in the prediction of the variance $\mathrm{V}_+$. Evaluating whether a control action reduces or increases the variance may thus be subject to uncertainty itself, which introduces numerical issues in downstream optimization techniques. The induced noise furthermore depends on the number of samples used to approximate the belief. Thus, numerical problems and approximation errors may only be avoided by choosing a high number of samples, rendering the underlying planning technique inefficient. 
In the following, we derive a different way of predicting the variance of object configurations without depending on the transition probability. In other words, we eliminate the need to sample from the belief transition distribution in \eqref{eq:transition} in order to predict the variance given a robot action. 

\begin{figure*}
    \centering
    \includegraphics[width=\linewidth]{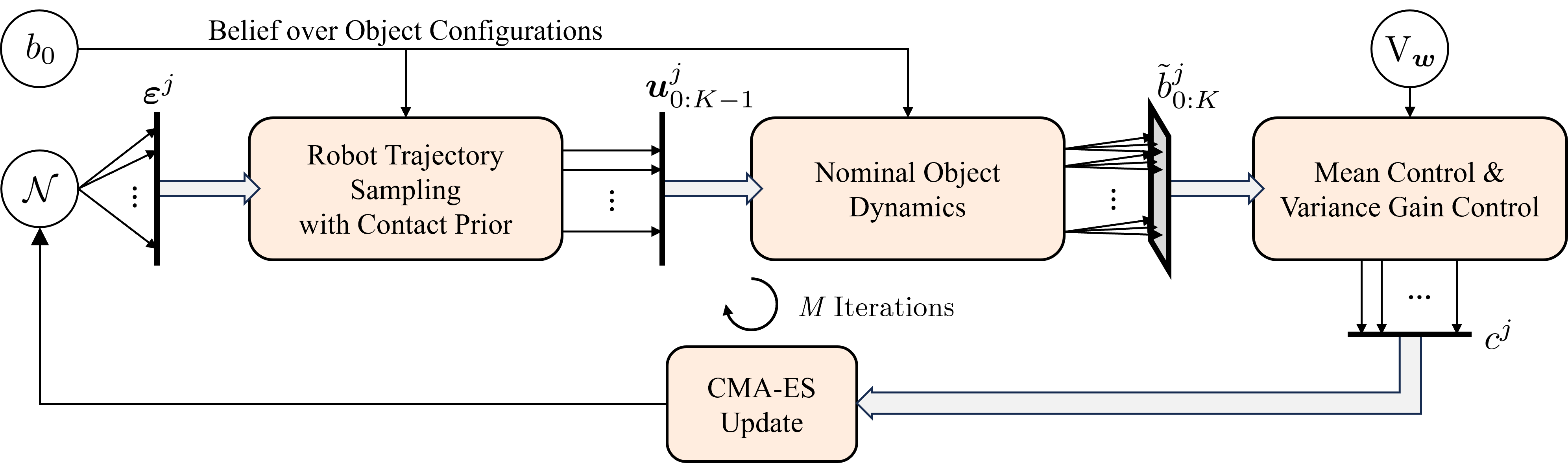}
    \caption{Block diagram depicting one iteration of BS-VP-STO. The algorithm starts with sampling a population of latent candidate trajectory variables $\bm{\varepsilon}$. These are then decoded into robot trajectories $\bm{q}^r_{0:K}$ using a contact prior. For each candidate trajectory the object belief is rolled-out using the nominal object dynamics. The variance gain together with the mean control cost is then used to compute the fitness of each candidate trajectory. Finally, the distribution of candidate trajectory variables weighted by the fitness is used to update the Gaussian approximation of the probability distribution using CMA-ES. After $M$ iterations, the algorithm returns the best performing candidate trajectory as solution.}
    \label{fig:bs-vp-sto}
\end{figure*}

Instead of predicting the updated variance after taking a control action by computing the variance of the predicted belief as in \eqref{eq:variance_pred_intractable}, we compute the variance of the noisy object dynamics in \eqref{eq:object_dynamics_noisy}, i.e.
\begin{align}
\label{eq:variance_pred_split}
\begin{split}
    \mathrm{V}_+ &= \mathrm{V}_{b, p_w}[\bm{f}(\bm{q}^o, \bm{u}) + \eta \bm{w}]\\
    &= \mathrm{V}_{b}[\bm{f}(\bm{q}^o, \bm{u})]  + \mathrm{V}_{b, p_w}[\eta \bm{w}] +\\
    2 &\mathrm{E}_{b, p_w}\left[(\bm{f}(\bm{q}^o, \bm{u}) - \mathrm{E}_{b}[\bm{f}(\bm{q}^o, \bm{u})])^\trsp (\eta \bm{w} - \mathrm{E}_{b, p_w}[\eta \bm{w}])\right].
\end{split}
\end{align}
At this point, we introduce the assumption that the expectation of the perturbation $\bm{w}$ is zero, i.e. $\mathrm{E}_{p_w}[\bm{w}] = \bm{0}$. As a result, perturbations are modeled as zero-mean noise in the tangential directions with respect to the contact point and zero noise in the normal direction of the contact. We furthermore note that the contact indicator $\eta$ and the perturbation are independent, such that the expectation of the product of the contact indicator and the perturbation is zero, i.e. $\mathrm{E}_{b, p_w}[\eta \bm{w}] = \mathrm{E}_{b}[\eta] \mathrm{E}_{p_w}[\bm{w}] = \bm{0}$. Consequently, the third term in \eqref{eq:variance_pred_split} simplifies to
\begin{align}
\label{eq:coavriance_simple}
\begin{split}
    &\mathrm{E}_{b, p_w}\left[(\bm{f}(\bm{q}^o, \bm{u}) - \mathrm{E}_{b}[\bm{f}(\bm{q}^o, \bm{u})])^\trsp (\eta \bm{w} - \mathrm{E}_{b, p_w}[\eta \bm{w}])\right] = \\
    &\mathrm{E}_{b, p_w}\left[\eta (\bm{f}(\bm{q}^o, \bm{u}) - \mathrm{E}_{b}[\bm{f}(\bm{q}^o, \bm{u})])^\trsp \bm{w}\right] = \\
    &\mathrm{E}_{b}\left[\eta (\bm{f}(\bm{q}^o, \bm{u}) - \mathrm{E}_{b}[\bm{f}(\bm{q}^o, \bm{u})])\right]^\trsp \mathrm{E}_{p_w}\left[\bm{w}\right] = \bm{0}.
\end{split}
\end{align}
As a result, the predicted variance in \eqref{eq:variance_pred_split} is equivalent to the sum of the nominally predicted variance and the variance of the induced noise. The predicted variance is thus
\begin{equation}
\label{eq:variance_pred_simple}
    \mathrm{V}_{+} = \mathrm{V}_{b}\left[\bm{f}(\bm{q}^o, \bm{u})\right] + \mathrm{V}_{b, p_w}\left[\eta \bm{w}\right].
\end{equation}
The left-hand term is computed through the nominal object dynamics $\bm{f}$, thus not including perturbations. The right-hand term is the variance contribution from the noise acting on the object when making contact. With the expectation of $p_w$ being zero, i.e. $\mathrm{E}_{b, p_w}[\eta \bm{w}] = \bm{0}$, the variance of the noise is given with
\begin{align}
\label{eq:variance_noise}
\begin{split}
    &\mathrm{V}_{b, p_w}\left[\eta \bm{w}\right] = \mathrm{E}_{b, p_w}\left[(\eta \bm{w})^\trsp (\eta \bm{w})\right] =\\ 
    &\mathrm{E}_{b, p_w}\left[\eta^2 \bm{w}^\trsp \bm{w}\right] = \mathrm{E}_{b}\left[\eta^2\right] \mathrm{E}_{p_w}\left[\bm{w}^\trsp \bm{w}\right].
\end{split}
\end{align}
With $\eta = \eta^2 \in \{0, 1\}$, the expectation of the squared contact indicator is equal to the expected contact indicator, i.e. $\mathrm{E}_{b}\left[\eta^2\right] = \mathrm{E}_{b}\left[\eta\right]$. The expected value of the contact indicator is the probability of the robot making contact with the object given a belief over object positions and a control action. Furthermore, note that due to the zero-mean property of our noise distribution, the expectation of the squared perturbation is equivalent to the variance of the perturbation, i.e. $\mathrm{E}_{p_w}\left[\bm{w}^\trsp \bm{w}\right] = \mathrm{V}_{\bm{w}}$. Inserting these equalities in \eqref{eq:variance_noise}, the variance of the applied perturbation can be expressed with
\begin{equation}
\label{eq:variance_noise_simple}
    \mathrm{V}_{b, p_w}\left[\eta \bm{w}\right] = \mathrm{E}_{b}\left[\eta\right] \mathrm{V}_{\bm{w}}.
\end{equation}
As a result of this section, we predict the variance of the object configuration with
\begin{equation}
\label{eq:variance_pred_result}
    \mathrm{V}_{+} = \mathrm{V}_{b}\left[\bm{f}(\bm{q}^o, \bm{u})\right] + \mathrm{E}_{b}\left[\eta\right] \mathrm{V}_{\bm{w}}.
\end{equation}
The result in \eqref{eq:variance_pred_result} is central to our contribution, as this allows us to predict the variance of the object configuration, i.e. the uncertainty, based on the variance of the perturbation $\mathrm{V}_{\bm{w}}$ that is assumed to be a constant value. Especially when those perturbations are constrained to the tangent space of the contact, sampling consistent perturbations involves expensive computations and makes the prediction of the variance stochastic.

\subsubsection{Monte-Carlo Approximation of the Nominal Dynamics.}\label{sec:particle}

We approximate the variance contribution resulting from the nominal contact dynamics using a non-parametric representation of the belief, i.e. particles.
We denote the approximated belief as a set of $N_p$ particles $\{{}^i\bm{q}^o, {}^i\alpha\}_{i=1}^{N_p}$, where each particle consists of a state sample ${}^i\bm{q}^o$ and a corresponding weight ${}^i\alpha$. The weight of a particle, $0 \leq {}^i\alpha \leq 1$, represents an approximate belief in the corresponding state sample ${}^i\bm{q}^o$. A prediction step consists of propagating each particle through the nominal contact dynamics, i.e.
\begin{equation}
    \label{eq:particle_prediction}
    {}^i\bm{q}^o_{+} = \bm{f}({}^i\bm{q}^o, \bm{u}), \, \forall i \in \{1, 2, .., N_p\}.
\end{equation}
Since we do not take any measurements during planning, we assume that the particle weights are constant and equally distributed, such that ${}^i\alpha = \sfrac{1}{N_p}, \; \forall i$. Based on the particle-representation of the belief, we estimate the variance of the object configuration with
\begin{equation}
    \label{eq:hat_variance}
    \hat{\mathrm{V}}_b[\bm{q}^o] = \frac{1}{N_p} \sum_{i=1}^{N_p} \left( {}^i\bm{q}^o - \bm{\mu}^o \right)^\trsp \left( {}^i\bm{q}^o - \bm{\mu}^o \right).
\end{equation}
The empirical mean object configuration is computed with $\bm{\mu}^o = \sfrac{1}{N_p} \sum_{i=1}^{N_p} {}^i\bm{q}^o$.
The probability of making contact $\mathrm{E}_{b}\left[\eta\right]$ is approximated as
\begin{equation}
    \label{eq:hat_eta}
    \hat{\mathrm{E}}_{b}\left[\eta\right] = \frac{1}{N_p} \sum_{i=1}^{N_p} {}^i\eta,
\end{equation}
where ${}^i\eta$ indicates if the $i$-th particle has an object configuration that is in contact with the robot. As a result, we approximate the overall variance prediction with
\begin{equation}
    \label{eq:variance_pred_result_approx}
    \hat{\mathrm{V}}_+ = \hat{\mathrm{V}}_b\left[\bm{f}(\bm{q}^o, \bm{u})\right] + \hat{\mathrm{E}}_{b}\left[\eta\right] \mathrm{V}_{\bm{w}}.
\end{equation}
When comparing the predicted variance $\hat{\mathrm{V}}_+$ against the variance of the current time step $\hat{\mathrm{V}}_b[\bm{q}^o]$, the same set of particles is used to compute the empirical variance as in \eqref{eq:hat_variance}. Thus, evaluating the approximated variance dynamics does not involve sampling and is thus deterministic. We exploit this approximation in Sec.~\ref{sec:bsvpsto} when optimizing robot trajectories based on the predicted variance.


\section{Stochastic Trajectory Optimization for Robust Manipulation}
\label{sec:bsvpsto}
Given our objective to push an object into a desired goal configuration subject to stochastic contact dynamics (cf.~\eqref{eq:problem_sep}), this section presents a framework that optimizes for robust robot trajectories directly in the belief space over possible object configurations. This framework extends our previous work \textit{VP-STO} (\cite{jankowski2023}) to belief-space via-point-based stochastic trajectory optimization (\textit{BS-VP-STO}). Our framework exploits the variance prediction developed in Sec.~\ref{sec:model} for synthesizing robust manipulation behavior. Due to the quasi-static pushing model in \eqref{eq:robot_dynamics} - \eqref{eq:penetration_constraint}, a robot trajectory is equivalent to a control trajectory, i.e. $\bm{q}^r_{1:K} = \bm{u}_{0:K-1}$. Thus, BS-VP-STO is a shooting method aiming at minimizing an objective that depends solely on the object configuration as in \eqref{eq:problem}.
Due to the non-smooth nature of the contact dynamics and the resulting non-smooth cost function with respect to the optimization variable, we approach the optimization problem with a gradient-free, i.e. zero-order, evolutionary optimization technique. 

Fig.~\ref{fig:bs-vp-sto} illustrates the optimization loop that is based on zero-order optimization of the variable $\bm{\varepsilon}$, which uniquely encodes a robot trajectory. At the beginning of the $m$-th iteration, we sample $N_{\mathrm{cand}}$ candidate trajectories from a latent Gaussian distribution that represents the current solution to the optimization problem:
\begin{equation}
    \bm{\varepsilon}^j \sim \mathcal{N}(\bar{\bm{\varepsilon}}_m, \bm{\Sigma}_m), \, \forall j \in \{1, 2, \hdots, N_{\mathrm{cand}} \}.
\end{equation}
The latent variable $\bm{\varepsilon}_j$ translates to a robot trajectory through an affine mapping $g$, i.e. $\bm{u}^j_{0:K-1}=g(\bm{\varepsilon}_j)$. This affine mapping imposes a contact prior on the sampling of robot trajectories, which is presented in Sec.~\ref{sec:sampling}. Given a candidate robot trajectory $\bm{u}^j_{0:K-1}$ and an initial belief over object positions $b_0 = p(\bm{q}^o)$, we compute the nominal belief dynamics. This results in a nominal belief trajectory $\tilde{b}^j_{0:K}$ for each candidate $\bm{\varepsilon}_j$. Based on the nominal belief trajectory we compute the step-wise predicted variance as developed in Sec.~\ref{sec:model}. The predicted variance is subsequently used in a cost and a constraint to the optimization problem, which is further outlined in Sec.~\ref{sec:robustness}. Together with a cost for controlling the mean of the object configuration, the total cost of each candidate trajectory is used to update the parameters of the latent Gaussian distribution, i.e. $\bar{\bm{\varepsilon}}_{m+1}, \bm{\Sigma}_{m+1}$ based on Covariance Matrix Adaptation (CMA-ES) (\cite{Hansen2016}).
After $M$ iterations, we return the best performing sample that returned the lowest cost.


\subsection{Variance Gain Control}
\label{sec:robustness}

\begin{figure}[t]
    \centering
    \includegraphics[width=\linewidth]{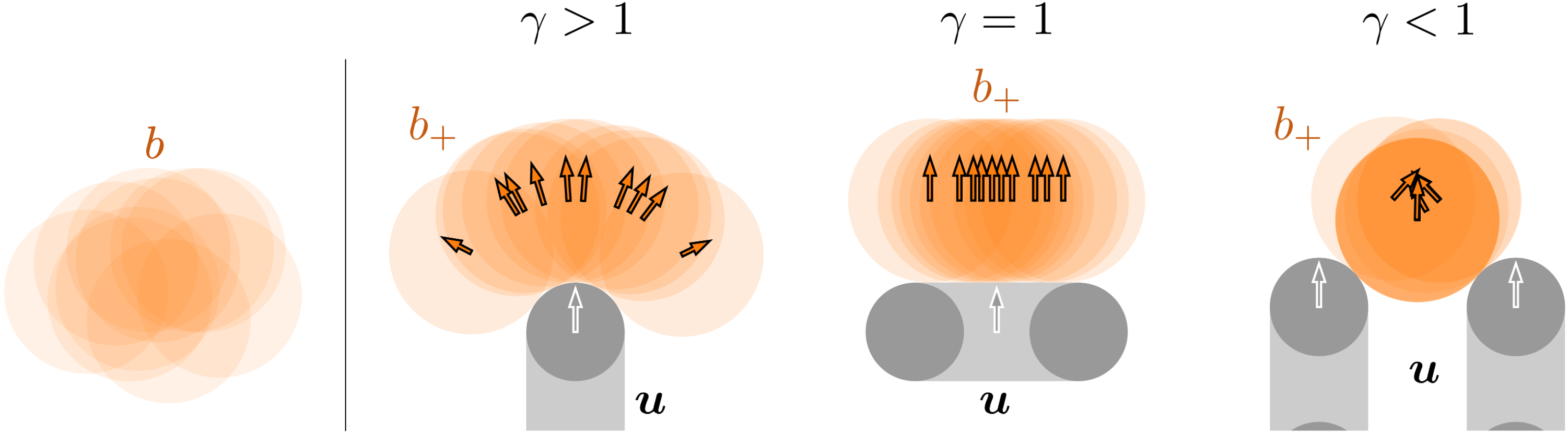}
    \caption{Belief dynamics through contact in a two-dimensional example. The three sub-figures on the right illustrate the predicted belief $b_+$ via samples in orange. All three cases started from the same initial belief $b$ that is depicted in the left-most sub-figure. The visualization of the prediction on the left shows an increase of the variance of the object position as a consequence of pushing with a single contact point ($\gamma > 1$). The second prediction shows a constant variance as a consequence of pushing with a flat contact surface ($\gamma = 1$). The right-most sub-figure shows a decrease of the variance of object position as a consequence of pushing with two contact points ($\gamma < 1$).}
    \label{fig:propagation}
\end{figure}

In Sec.~\ref{sec:model} we derive an approximation of the one-step prediction of the variance of the object configuration. However, this does not enable the prediction of the variance after multiple time-steps, which is due to the fact that the prediction of the variance $\hat{\mathrm{V}}_{k+1}$ in \eqref{eq:variance_pred_result_approx} requires the belief of the previous time step $b_k$ to be known. Therefore, instead of directly controlling the variance at the end of the trajectory $\hat{\mathrm{V}}_{K}$, we propose to control the predicted variance at each time step. Given a robot trajectory $\bm{u}^j_{0:K-1}$, we thus compute the nominal belief over object configurations at each time step, i.e. $\tilde{b}_{k}$, via the the nominal object dynamics in \eqref{eq:object_dynamics}. Given the particle set that approximates the initial belief $b_0 = p(\bm{q}^o)$ as an input to BS-VP-STO, the nominal belief rollout is computed by applying the nominal forward dynamics to all particles:
\begin{equation}
    \label{eq:particle_rollout}
    {}^i\bm{q}^o_{k+1} = \bm{f}\left({}^i\bm{q}^o_{k}, \bm{u}_k\right), \, \forall i \in \{1, 2, .., N_p\}.
\end{equation}
The one-step prediction of the variance at each time step can then be computed as
\begin{align}
\begin{split}
    \hat{\mathrm{V}}_{k+1} &= \hat{\mathrm{V}}_{\tilde{b}_{k}}\left[\bm{f}(\bm{q}^o, \bm{u}_k)\right] + \hat{\mathrm{E}}_{\tilde{b}_{k}}\left[\eta\right] \mathrm{V}_{\bm{w}}\\
    &= \hat{\mathrm{V}}_{\tilde{b}_{k+1}}\left[\bm{q}^o\right] + \hat{\mathrm{E}}_{\tilde{b}_{k}}\left[\eta\right] \mathrm{V}_{\bm{w}}.
\end{split}
\end{align}
In order to quantify the change of uncertainty due to a given control action $\bm{u}_k$, we are interested in the amount of variance that is gained over one time step. Note that the variance of a continuous random variable is closely related to its differential entropy. While the variance as defined in \eqref{eq:object_variance} is equivalent to the trace of the covariance matrix, i.e. the sum over all eigenvalues, the upper bound of the differential entropy is monotonic in the determinant of the covariance matrix, i.e. the product over all eigenvalues (\cite{Cover05}). As a result, the variance of a continuous random variable yields an upper bound for its entropy.
Therefore, we introduce a new metric $\gamma$, which we call \textit{variance gain}, measuring the relative change of the variance after applying an action $\bm{u}_k$, i.e.
\begin{equation}
\label{eq:variance_gain}
    \gamma_k = \frac{\hat{\mathrm{V}}_{k+1}^+}{\hat{\mathrm{V}}_{\tilde{b}_{k}}\left[\bm{q}^o\right] + \mathrm{V}_{\bm{w}}}
    = \frac{\hat{\mathrm{V}}_{\tilde{b}_{k+1}}\left[\bm{q}^o\right] + \hat{\mathrm{E}}_{\tilde{b}_{k}}\left[\eta\right] \mathrm{V}_{\bm{w}}}{\hat{\mathrm{V}}_{\tilde{b}_{k}}\left[\bm{q}^o\right] + \mathrm{V}_{\bm{w}}}.
\end{equation}
The variance gain $\gamma$ is the ratio of the output variance, i.e. the predicted variance $\hat{\mathrm{V}}_{k+1}$, to the input variance, i.e. $\hat{\mathrm{V}}_{k} + \mathrm{V}_{\bm{w}}$. Fig.~\ref{fig:propagation} illustrates three different robot actions resulting in different variance gains. It shows that the contact geometry plays a crucial role when planning to make contact between the robot and an object. The variance gain $\gamma$ reflects that the contact geometry affects the robustness of a contact, e.g. when pushing. The left sub-figure shows that using one finger for pushing a circular object with uncertain location results in an increase of variance ($\gamma > 1$). In contrast, the right sub-figure shows that using two fingers with one finger pushing the object towards the other finger results in a decrease of variance ($\gamma < 1$). Using a flat contact surface to push an object with uncertain location keeps the variance constant as shown in the middle sub-figure ($\gamma = 1$). Note that the variance gain is lower or equal to one for robot actions that have zero probability of making contact with the object. In this case, the belief does not change, i.e. $\hat{\mathrm{V}}_{\tilde{b}_{k+1}}\left[\bm{q}^o\right] = \hat{\mathrm{V}}_{\tilde{b}_{k}}\left[\bm{q}^o\right]$, since no perturbation is injected into the belief, i.e. $\hat{\mathrm{E}}_{\tilde{b}_{k}}\left[\eta\right] \mathrm{V}_{\bm{w}} = 0$. Thus, a no-contact action results in
\begin{equation}
    \gamma_{\mathrm{no-contact}} = \frac{\hat{\mathrm{V}}_{\tilde{b}_{k}}\left[\bm{q}^o\right]}{\hat{\mathrm{V}}_{\tilde{b}_{k}}\left[\bm{q}^o\right] + \mathrm{V}_{\bm{w}}} \leq 1.
\end{equation}

We propose to enforce robustness in the optimization problem by constraining the solution to variance gains smaller or equal to one at all steps, i.e.
\begin{equation}
    \label{eq:robustness_constraint}
    \gamma_k \leq 1 \quad \forall \, k.
\end{equation}
Due to the zero-order technique that we use for optimizing the robot trajectory, we deploy the robustness constraint as a barrier cost, i.e. a discontinuous cost that is zero if the constraint is satisfied and returns a high value if the constraint is violated. We can further reduce the predicted variance by adding a cost term that is active if the constraint is already satisfied. This is encapsulated in the following cost term
\begin{equation}
    \label{eq:robustness_cost}
    c_{\mathrm{robust}} = \lambda_c \prod_{k=0}^{K-1} e^{-\frac{1 - \gamma_k}{K-1}},
\end{equation}
with a discontinuous cost weight
\begin{equation}
\label{eq:robustness_cost_weight}
    \lambda_c = \begin{cases}
        1 \quad \quad \mathrm{if} \max_k \gamma_k \leq 1, \\
        10^3 \hfill \mathrm{else}.
    \end{cases}
\end{equation}
The total cost of a candidate robot trajectory is computed as the sum of the variance gain control cost in \eqref{eq:robustness_cost} and a task-specific cost $c_{\mathrm{task}}$ that computes the deviation of the mean of the object configuration to the desired goal configuration.

\subsection{Trajectory Sampling with a Contact Prior}\label{sec:sampling}

\begin{figure}
    \centering
    \includegraphics[width=\linewidth]{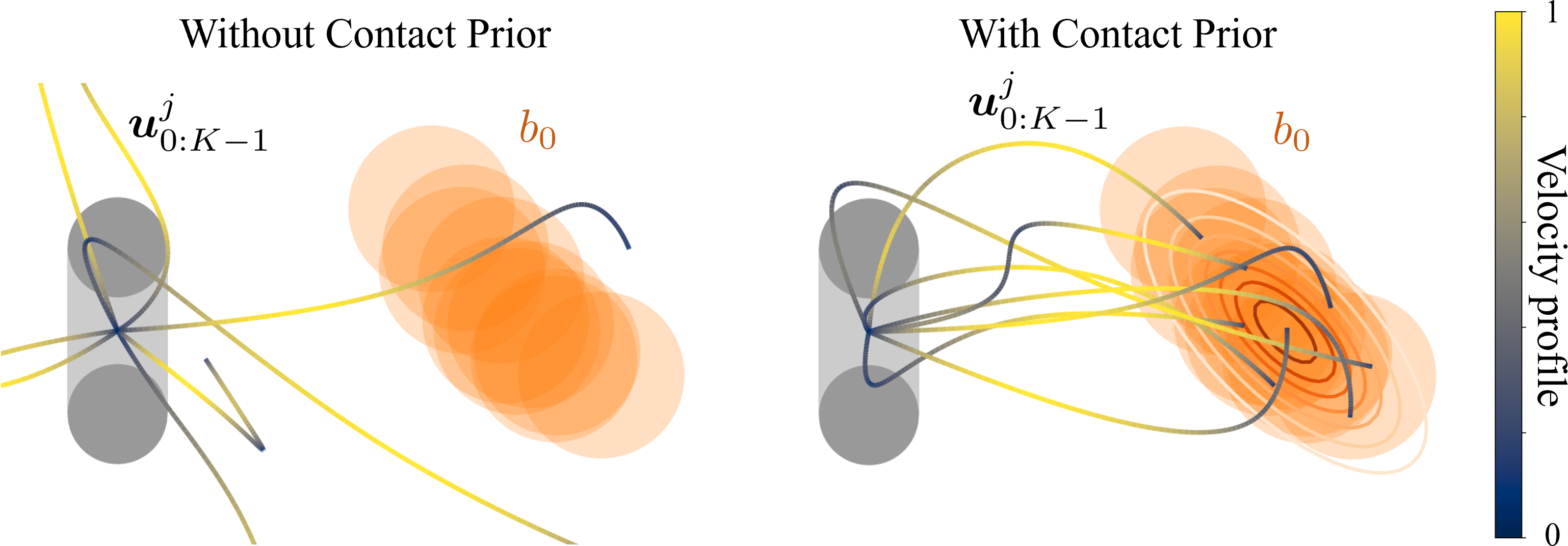}
    \caption{Comparison of uninformed and informed sampling of robot trajectories. The left sub-figure shows trajectory samples drawn from a probability distribution computed without a contact prior ($\bm{Q}_{\bm{q}} = \bm{0}$). The right sub-figure shows trajectory samples drawn from a probability distribution computed with a contact prior ($\bm{Q}_{\bm{q}} > \bm{0}$). The contact prior guides the sampling of robot trajectories towards regions where the robot is likely to make contact with the object given the object belief.}
    \label{fig:candidate-sampling}
\end{figure}

The efficiency of sampling-based optimization algorithms depend on the quality of the generated samples. In this section, we present how the object belief can be used to inform the sampling of robot trajectories for manipulation tasks. In general, it is desirable to sample trajectories with a likelihood that is proportional to the negative cost that can be expected from executing the trajectory. In the following, we denote a robot trajectory with $\bm{q}^r_{0:K} = \left[ \bm{q}^r_{0}, \bm{q}^r_{1}, \hdots ,\bm{q}^r_{K} \right]$, where $K$ corresponds to the number of discretized steps captured by the trajectory. Suppose that the cost is given as a function of the robot trajectory, i.e. $c = f_c(\bm{q}^r_{0:K})$, then we would like to sample robot trajectories from a corresponding probability distribution with
\begin{equation}
    p(\bm{q}^r_{0:K}) \propto \exp\left(-f_c(\bm{q}^r_{0:K})\right).
\end{equation}
While we do not have access to such a generative probability distribution, we approximate it with a prior that improves the sample-efficiency of the optimization algorithm compared to sampling initial guesses from an uninformed probability distribution. The idea is to sample robot trajectories in regions where the robot is likely to make contact with the object given its belief. This is motivated by the observation that making contact is a necessary precondition of manipulating an object.

Figure~\ref{fig:candidate-sampling} illustrates the impact of the contact prior on trajectory samples. Without the contact prior, a large portion of the candidate trajectories explores regions in which the robot does not move into the object belief, and thus does not contribute to the optimization process.

\subsubsection{Via-point-based Trajectory Parameterization.}

In order to efficiently synthesize robot trajectories in a low-dimensional space, we adopt the via-point-based trajectory representation as in \cite{jankowski2023}. The robot configuration is given with
\begin{equation}
\label{eq:q_obf}
    \bm{q}^r(t) = \bm{\Phi}_{\mathrm{via}}(t) \bm{\theta} + \bm{\phi}_{0}(t, \bm{q}_0^r, \dot{\bm{q}}_{0}^r),
\end{equation}
where the robot trajectory is parameterized with
\begin{equation}
\label{eq:theta}
    \bm{\theta} = \begin{pmatrix} \bm{q}^1_{\mathrm{via}} \\ \vdots \\ \bm{q}_{\mathrm{via}}^{N} \end{pmatrix} \in \mathbb{R}^{N \cdot n_{\mathrm{dof}}^r}.
\end{equation}
The trajectory parameter $\bm{\theta}$ contains $N$ via-points the trajectory passes through. The basis functions $\bm{\Phi}_{\mathrm{via}}(t)$ enforce that the trajectory passes exactly through the via configurations while smoothly interpolating with minimal acceleration. The basis functions furthermore enforce that the velocity at the end of the trajectory is zero. Note that the last $n_{\mathrm{dof}}^r$ elements of $\bm{\theta}$ are the final robot configuration at the end of the trajectory, i.e. $\bm{q}^r(T) = \bm{q}_{K}^r = \bm{q}_{\mathrm{via}}^{N}$. The basis offset $\bm{\phi}_{0}(t, \bm{q}_0^r, \dot{\bm{q}}_{0}^r)$ incorporates the initial robot position $\bm{q}_{0}^r$ and velocity with $\dot{\bm{q}}_0^r$. $T$ denotes the duration of the trajectory. We use the time scaling algorithm in \cite{jankowski2023} for computing the duration of a trajectory based on a given parameter $\bm{\theta}$ such that user-defined velocity and acceleration limits are enforced. For implementation details on how to compute the basis functions and offsets, please refer to \cite{jankowski2022}.

In the following, we are interested in computing a Gaussian distribution of the via-points $\bm{\theta}$ to efficiently sample from. Due to the affine mapping from via-points to robot trajectories in \eqref{eq:q_obf}, this corresponds to sampling from a Gaussian distribution of continuous robot trajectories.

\subsubsection{Gaussian Contact Prior.}

The contact prior is a probability distribution that guides the sampling of robot trajectories towards regions where the robot is likely to make contact with the object. To allow for variations in how to approach the contact with the object, we only consider the final robot configuration of the trajectory to be subject to the contact prior. This is incorporated by exploiting the parameterization of the robot trajectory in \eqref{eq:q_obf}, where the final robot configuration is explicitly given by the last $n_{\mathrm{dof}}^r$ elements of $\bm{\theta}$.

Suppose that a conditional Gaussian distribution
\begin{equation}
    p_{\mathrm{c}}(\bm{q}^r | \bm{q}^o) = \mathcal{N}(\bm{f}_{\mathrm{c}}(\bm{q}^o), \bm{\Sigma}^{r|o})
\end{equation}
approximates the probability density of a robot configuration making contact with the object. Furthermore, suppose that the object configuration is Gaussian distributed as well with $\bm{q}^o \sim \mathcal{N}(\bm{\mu}^o, \bm{\Sigma}^o)$. Practically, we find the Gaussian distribution of object configurations by approximating the initial belief $b_0$ with a Gaussian distribution for computing the contact prior. This lets us compute a probability distribution over robot configurations indicating how likely it is to establish a contact between the robot and the object:
\begin{equation}
\label{eq:contact_prior}
    p_{\mathrm{c}}(\bm{q}^r) = \mathcal{N}(\bm{f}_{\mathrm{c}}(\bm{\mu}^o), \bm{\Sigma}^{r|o} + \bm{A} \bm{\Sigma}^o \bm{A}^\trsp),
\end{equation}
with $\bm{A} = \sfrac{\partial \bm{f}_{\mathrm{c}}}{\partial \bm{q}^o} |_{\bm{\mu}^o}$. The corresponding prior on the trajectory parameter $\bm{\theta}$ is then given by
\begin{equation}
\label{eq:contact_prior_via}
    p_{\mathrm{c}}\left(\bm{\theta}{=}\begin{pmatrix} \bm{q}^1_{\mathrm{via}} \\ \vdots \\ \bm{q}_{\mathrm{via}}^{N} \end{pmatrix}\right) = 
    \mathcal{N}\left( \begin{pmatrix} \bm{0} \\ \vdots \\ \bar{\bm{q}}_{\mathrm{c}} \end{pmatrix},
    \begin{pmatrix} \bm{0} & \! \cdots & \! \! \bm{0} \\
                    \vdots & \! \ddots & \! \! \vdots \\
                    \bm{0} & \! \cdots & \! \! \bm{Q}_{\bm{q}} \end{pmatrix}^{-1} \right),
\end{equation}
where $\bm{Q}_{\bm{q}} = \left(\bm{\Sigma}^{r|o} + \bm{A} \bm{\Sigma}^o \bm{A}^\trsp\right)^{-1}$ describes the precision matrix of the contact prior with respect to the mean contact configuration $\bar{\bm{q}}_{\mathrm{c}} = \bm{f}_{\mathrm{c}}(\bm{\mu}^o)$. We denote the contact prior with
\begin{equation}
    p_{\mathrm{c}}(\bm{\theta}) = \mathcal{N}(\bar{\bm{\theta}}_\mathrm{c}, \bm{Q}_{\bm{\theta}}^{-1}).
\end{equation}
Note that the resulting covariance matrix $\bm{Q}_{\bm{\theta}}^{-1}$ is degenerated due to zero-precision values for the via-points except for $\bm{q}_{\mathrm{via}}^{N}$.
For resolving the degeneration, we regularize the covariance matrix by combining the contact prior with a smoothness prior as described in the following.

\subsubsection{Gaussian Contact Prior in Joint Space.}\label{sec:contact_prior_joint}

Optimizing in the joint space of an articulated robot such as robot arms may be beneficial when kinematic and dynamic limitations are to be considered during planning. Sampling in joint space requires to represent the contact prior in joint space as well. Suppose that the robot's end-effector, that is supposed to manipulate the object, has a configuration given by $\bm{x}^r$ which is computed from the robot's joint positions $\bm{q}^r$ via forward kinematics $\bm{x}^r = \bm{f}_{\mathrm{fk}}(\bm{q}^r)$. We may adopt the contact prior in \eqref{eq:contact_prior} to formulate a prior distribution over configurations of the end-effector, i.e.
\begin{equation}
\label{eq:contact_prior_EE}
    p_{\mathrm{c}}(\bm{x}^r) = \mathcal{N}(\bm{f}_{\mathrm{c}}(\bm{\mu}^o), \bm{\Sigma}^{r|o} + \bm{A} \bm{\Sigma}^o \bm{A}^\trsp).
\end{equation}
For computing a corresponding Gaussian distribution in joint space, the forward kinematics are linearized around a mean joint position $\bar{\bm{q}}_{\mathrm{c}}^r$ with
\begin{equation}
    \bm{x}^r \approx \bm{f}_{\mathrm{fk}}(\bar{\bm{q}}_{\mathrm{c}}^r) + \bm{J}(\bar{\bm{q}}_{\mathrm{c}}^r) \left( \bm{q}^r - \bar{\bm{q}}_{\mathrm{c}}^r\right),
\end{equation}
where $\bm{J}(\bm{q}) = \sfrac{\partial \bm{x}^r}{\partial \bm{q}^r}|_{\bar{\bm{q}}_{\mathrm{c}}^r}$ denotes the Jacobian with respect to the end-effector configuration. The mean joint position $\bar{\bm{q}}_{\mathrm{c}}^r$ can be computed via inverse kinematics with respect to the mean end-effector configuration $\bm{f}_{\mathrm{c}}(\bm{\mu}^o)$. Consequently, the Gaussian contact prior for the end-effector can be locally transformed into the joint space, resulting in a Gaussian distribution $p_{\mathrm{c}}(\bm{q}^r) = \mathcal{N}(\bar{\bm{q}}_{\mathrm{c}}^r, \bm{Q}_{\bm{q}}^{-1})$. The joint space contact precision matrix is computed with
\begin{equation}
    \bm{Q}_{\bm{q}} = \bm{J}(\bar{\bm{q}}_{\mathrm{c}}^r)^\trsp \left( \bm{\Sigma}^{r|o} + \bm{A} \bm{\Sigma}^o \bm{A}^\trsp \right)^{-1} \bm{J}(\bar{\bm{q}}_{\mathrm{c}}^r).
\end{equation}

\subsubsection{Gaussian Smoothness Prior.}

\begin{figure}
    \centering
    \includegraphics[width=\linewidth]{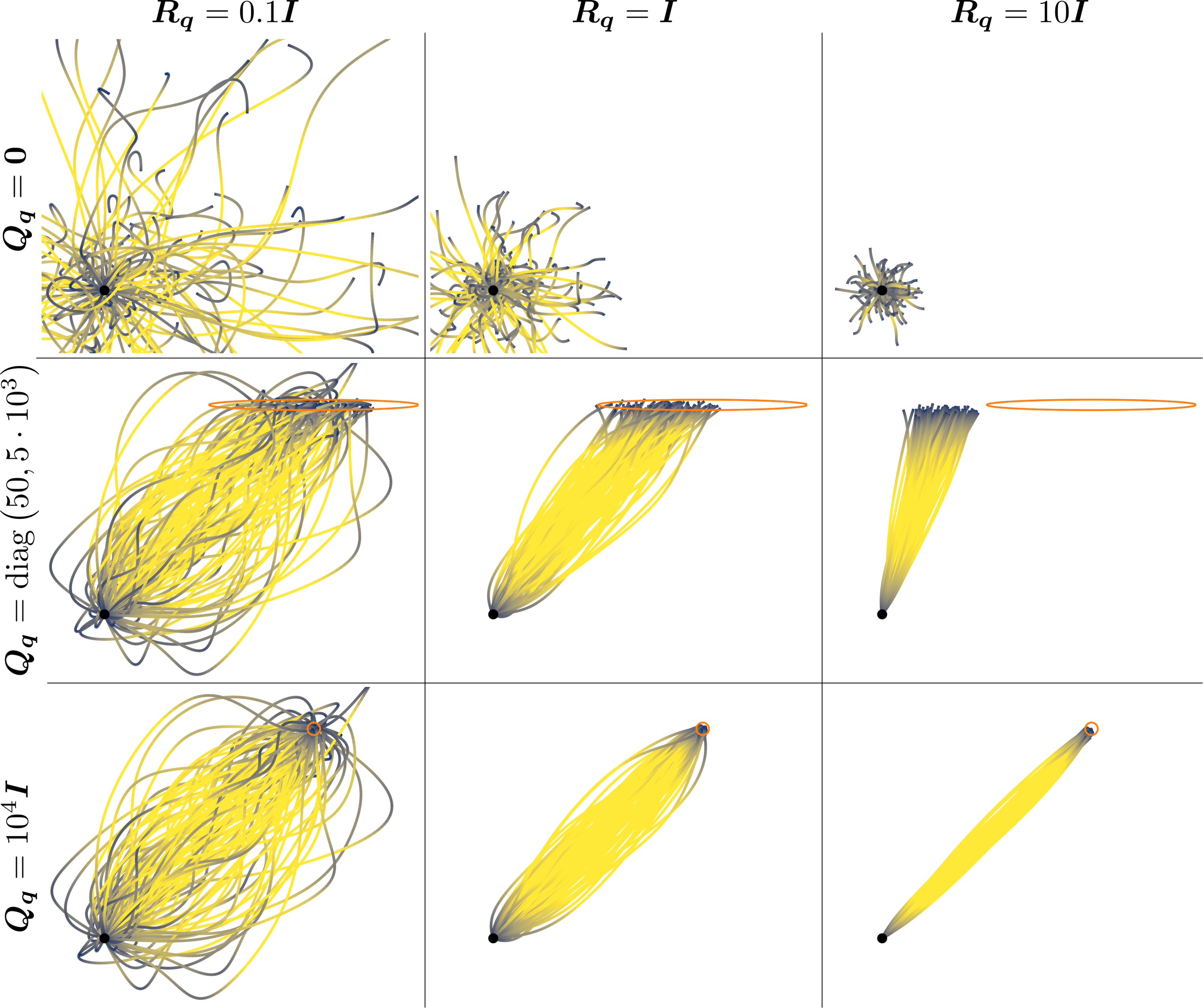}
    \caption{Smooth trajectories sampled from the product of the smoothness prior and the contact prior. The contact prior is indicated by the orange ellipses and circles with the mean of the contact prior in the center. The velocity profile of the trajectories is encoded through color with low velocities in blue and high velocities in yellow. Note that all trajectories start and end with exactly zero velocity.}
    \label{fig:QvsR}
\end{figure}

The smoothness prior, introduced in our previous work (\cite{jankowski2023}), incorporates temporal correlations between via-points by computing a Gaussian distribution that expresses a high likelihood for low-acceleration profiles. A typical objective in trajectory optimization is to minimize the integral over squared accelerations of the candidate trajectory, i.e.
\begin{equation}
\label{eq:smoothness_cost_simple}
    J_{\mathrm{s}} = \frac{1}{2} \int_0^T \ddot{\bm{q}}^{r \trsp}(t) \bm{R}_{\bm{q}} \ddot{\bm{q}}^r(t) dt.
\end{equation}
The positive definite matrix $\bm{R}_{\bm{q}}$ encodes the desired smoothing for the individual degrees of freedom. Using the parameterization in \eqref{eq:q_obf}, this objective can be expressed using the via-point parameter $\bm{\theta}$ and the initial conditions for the trajectory $\bm{q}_0^r, \dot{\bm{q}}_{0}^r$, i.e. $J_{\mathrm{s}}(\bm{\theta}, \bm{q}_0^r, \dot{\bm{q}}_{0}^r)$.
As a next step, we express the smoothness prior as a probability distribution parameterized with the negative objective in \eqref{eq:smoothness_cost_simple} with
\begin{equation}
\label{eq:cost2prob}
    p_\mathrm{s}(\bm{\theta}, \bm{q}_0^r, \dot{\bm{q}}_{0}^r) \propto e^{- J_{\mathrm{s}}(\bm{\theta}, \bm{q}_0^r, \dot{\bm{q}}_{0}^r)}.
\end{equation}
Interestingly, this results in a joint Gaussian distribution over the trajectory parameter and the initial conditions. We then compute a Gaussian smoothness prior on the trajectory parameter by conditioning on the initial conditions, i.e.
\begin{equation}
    p_\mathrm{s}(\bm{\theta} | \bm{q}_0^r, \dot{\bm{q}}_{0}^r) = \mathcal{N}(\bar{\bm{\theta}}_\mathrm{s}, \bm{R}_{\bm{\theta}}^{-1}),
\end{equation}
with the precision matrix being computed with
\begin{equation}
\label{eq:smoothness_prec}
    \bm{R}_{\bm{\theta}} = \int_0^T \ddot{\bm{\Phi}}^\trsp_{\mathrm{via}}(t) \bm{R}_{\bm{q}} \ddot{\bm{\Phi}}_{\mathrm{via}}(t) dt.
\end{equation}
Please refer to the appendix for the derivation of \eqref{eq:smoothness_prec} and for details on how to compute the smoothness prior mean $\bar{\bm{\theta}}_\mathrm{s}$. Note that the smoothness precision matrix $\bm{R}_{\bm{\theta}}$ can be computed offline as it does not depend on the trajectory parameter $\bm{\theta}$.

\subsubsection{Product of Gaussian Priors.}

\begin{figure*}[t]
    \centering
    \includegraphics[width=\linewidth]{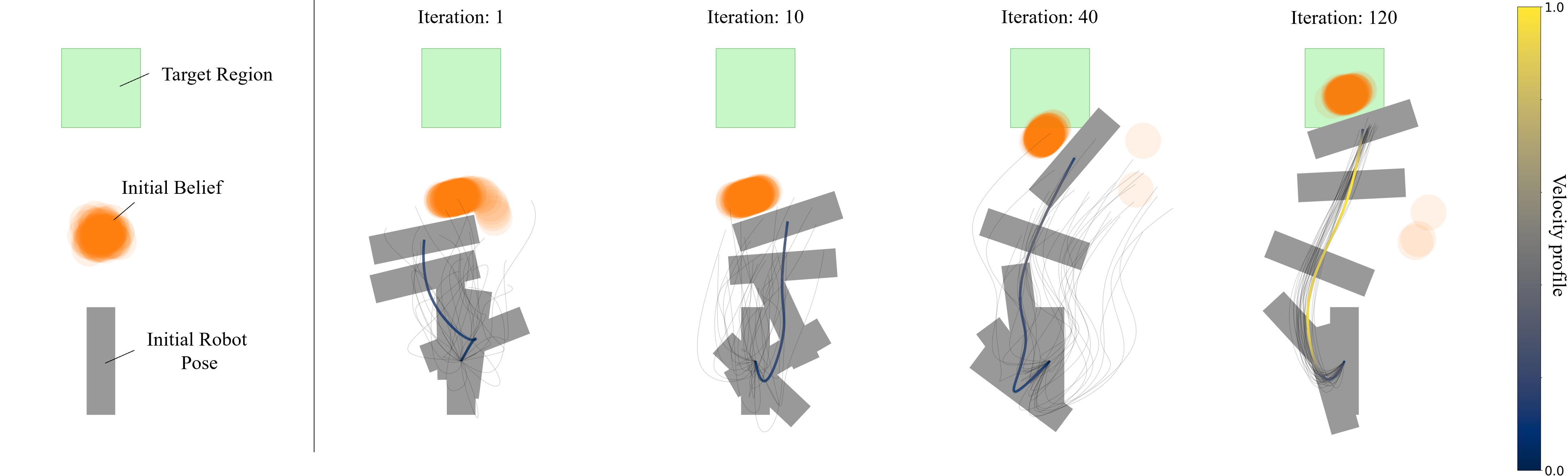}
    \caption{BS-VP-STO optimizing the trajectory of a rectangular robot to push a circular object into a goal region subject to uncertain initial object location (represented by a particle-based belief representation) and uncertain contact dynamics. The sub-figures show the best performing candidate solution with the corresponding velocity profile, as well as the other trajectory samples in light gray, after 1, 10, 40 and 120 iterations from left to right.}
    \label{fig:exp_sim}
\end{figure*}

Given the two priors for making contact and smooth trajectories respectively, we find the informed via-point distribution by fusing the two probabilistic priors via computing the normalized product of the two priors, such that
\begin{equation}
    p(\bm{\theta}) = \mathcal{N}(\bar{\bm{\theta}}, \bm{\Sigma}_{\bm{\theta}}) \propto p_{\mathrm{c}}(\bm{\theta}) p_\mathrm{s}(\bm{\theta} | \bm{q}_0^r, \dot{\bm{q}}_{0}^r).
\end{equation}
The product of two multivariate Gaussians is again a multivariate Gaussian, with the resulting parameters given by
\begin{subequations}
    \label{eq:prior_product}
    \begin{equation}
    \bm{\Sigma}_{\bm{\theta}} = \left(\bm{Q}_{\bm{\theta}} + \bm{R}_{\bm{\theta}} \right)^{-1},
    \end{equation}
    \begin{equation}
    \bar{\bm{\theta}} = \bm{\Sigma}_{\bm{\theta}} (\bm{Q}_{\bm{\theta}} \bar{\bm{\theta}}_\mathrm{c} + \bm{R}_{\bm{\theta}} \bar{\bm{\theta}}_\mathrm{s}).
    \end{equation}
\end{subequations}

Fig.~\ref{fig:QvsR} illustrates trajectories sampled from the product of the contact prior, parameterized by $\bm{Q}_{\bm{q}}$, and the smoothness prior, parameterized by $\bm{R}_{\bm{q}}$. Within each sub-figure, all trajectories are drawn from a single Gaussian distribution. Note that all trajectories start and end with zero velocity.

\subsubsection{Optimizing and Sampling in Latent Space.}

Given the product of priors, we use the informed generative via-point distribution $p(\bm{\theta})$ as a probabilistic initial guess for optimizing robot trajectories with CMA-ES.
Instead of directly sampling trajectory candidates $\bm{\theta}$ from an uninformed distribution, e.g. white noise, we sample and optimize for $\bm{\varepsilon} \in \mathbb{R}^{N n_{\mathrm{dof}}^r}$. For a given $\bm{\varepsilon}$, we compute $\bm{\theta}$ through an affine transformation as follows:
\begin{equation}
\label{eq:affine_sampling}
    \bm{\theta} = \bar{\bm{\theta}} + \bm{L}_{\bm{\theta}} \bm{\varepsilon}.
\end{equation}
Here, $\bm{L}_{\bm{\theta}}$ is the Cholesky decomposition of the covariance matrix $\bm{\Sigma}_{\bm{\theta}}$. The parameters $\bar{\bm{\theta}}$ and $\bm{\Sigma}_{\bm{\theta}}$ incorporate the prior as defined in \eqref{eq:prior_product}. The idea of this additional transformation is to decouple the optimization variable $\bm{\varepsilon}$ from the particular prior. In each iteration $m$ of BS-VP-STO, we obtain the new population of candidate solutions by sampling $N_{\mathrm{cand}}$ robot trajectories via 
\begin{equation}
\label{eq:affine_sampling_dist}
    \bm{\theta} \sim \mathcal{N}\left(\bar{\bm{\theta}} + \bm{L}_{\bm{\theta}} \bar{\bm{\varepsilon}}_m, \bm{L}_{\bm{\theta}} \bm{\Sigma}_m \bm{L}_{\bm{\theta}}^\trsp\right).
\end{equation}
When initializing the CMA-ES distribution as white noise, i.e. $\bar{\bm{\varepsilon}}_0 = \bm{0}$ and $\bm{\Sigma}_0 = \bm{I}$, we effectively sample the first population from the informed distribution in \eqref{eq:prior_product}, as inserting the initial parameters into \eqref{eq:affine_sampling_dist} yields
\begin{equation}
\label{eq:initial_dist}
    \bm{\theta} \sim \mathcal{N}\left(\bar{\bm{\theta}}, \bm{L}_{\bm{\theta}} \bm{L}_{\bm{\theta}}^\trsp\right) = \mathcal{N}\left(\bar{\bm{\theta}}, \bm{\Sigma}_{\bm{\theta}}\right).
\end{equation}
Eventually, given a sampled trajectory parameter $\bm{\theta}$, we find the control trajectory with respect to the system in \eqref{eq:object_dynamics} by discretizing the robot trajectory in \eqref{eq:q_obf}, i.e.
\begin{equation}
    \bm{u}_k = \bm{q}^r\left(t = T \cdot \frac{k+1}{K}\right).
\end{equation}
Note that due to the quasi-static model in \eqref{eq:object_dynamics}, the dynamics can be rolled out with an arbitrary temporal resolution.\newline

\noindent
\textbf{Example 2. (Single-horizon Robust 2D Object-Pushing).} We showcase the BS-VP-STO pipeline over multiple iterations for a 2D object pushing example, illustrated in Fig.~\ref{fig:exp_sim}. The task for the robot, a rectangular geometry, is to push the object, a circular geometry, into a target region. We consider two sources of uncertainty in this example: a) The initial position of the object is uncertain, which is reflected by an initial belief; and b) the contact dynamics are uncertain. This task requires exploring contact modes that are robust to these uncertainties, as implicitly done by the presented planning algorithm.

After the first iteration, the best solution corresponds to the robot making contact with its long side. Note that this solution corresponds to the best candidate of the initial population sampled from the product of Gaussian priors, without any CMA-ES updates. Due to the probabilistic contact prior, almost all of the 30 initial candidates bring the robot into contact with the object, enabling an informative sampling of the cost landscape. After 10 iterations of BS-VP-STO, the algorithm found a solution making robust contact with the object while moving it slightly towards the target area. The solution after 40 iterations enables the robot to almost push the object into the target area. After 120 iterations, the algorithm found a solution for pushing the object robustly into the target area, while also optimizing the overall motion duration which is incorporated into  $c_{\mathrm{task}}$. The tight distribution of candidate solutions after 120 iterations indicates that CMA-ES has converged.


\section{Receding-horizon BS-VP-STO}\label{sec:multi_bsvpsto}

Planning a pushing maneuver over a single long horizon is challenging for two reasons: \textit{i)} The dimensionality of the solution space of the optimization problem grows as the solution requires higher expressiveness. In the presented algorithm, this is reflected by an increasing number of via-points $N$ that parameterize the robot trajectory. \textit{ii)} In BS-VP-STO, the belief is rolled out using the nominal object dynamics, i.e. without inducing noise. For robust candidate trajectories, the belief may collapse to a Dirac-delta distribution after $k_{\mathrm{collapse}}$ steps, i.e. $\tilde{b}_k = \delta(\bm{q}^o_k), \forall k \geq k_{\mathrm{collapse}}$, resulting in zero variance. In this case, the variance gain at those time steps is either $\gamma_{k} = 0$ if the robot does not touch the object, or $\gamma_{k} = 1$ if the robot touches the object. Thus, the optimization is strongly biased towards not touching the object after $k_{\mathrm{collapse}}$ steps.

\RestyleAlgo{ruled}
\SetKwComment{Comment}{/* }{ */}
\SetKw{AND}{and}
\SetKw{IN}{Input:}

\begin{algorithm}[t!]
\DontPrintSemicolon
\caption{Receding-horizon BS-VP-STO}\label{alg}
\KwIn{Robot configuration $\bm{q}^r_0$ and velocity $\dot{\bm{q}}^r_0$, \hspace{1pt} object belief $b_0$, receding horizon length $H$.}
\KwOut{Robot trajectory $\bm{u}$.}
$\bm{u} \gets \emptyset$\;
\While{task not solved}{
    $\bm{q}^{r*}_{0:K}, \dot{\bm{q}}^{r*}_{0:K} \gets$ BS-VP-STO$(\bm{q}^r_0, \dot{\bm{q}}^r_0, b_0)$\;
    $\bm{u}^*_{0:H-1} \gets \bm{q}^{r*}_{1:H}$\;
    \For{$k \gets 0$ to $H-1$}{
    \tcp{Stochastic rollout}
        ${}^i\bm{q}^o_{k+1} \sim p(\cdot | {}^i\bm{q}^o_k, \bm{u}^*_k), \, \forall i \in \{1, 2, .., N_p\}$\;
    }
    $\bm{q}^r_0, \dot{\bm{q}}^{r}_{0} \gets \bm{q}^{r*}_{H}, \dot{\bm{q}}^{r*}_{H}$\;
    $b_0 \gets \{{}^i\bm{q}^o_H\}_{i=1}^{N_p}$\;
    $\bm{u} \gets$ concatenate$\left(\bm{u}, \bm{u}^*_{0:H-1}\right)$\;
}
\end{algorithm}

For these reasons, we propose \textit{receding-horizon} BS-VP-STO, a planning scheme for pushing maneuvers over longer horizons. The scheme alternates between computing a robust push via BS-VP-STO and performing a stochastic rollout of the solution.
This allows to plan pushing maneuvers over multiple shorter horizons, while optimizing for task progress, i.e. pushing the object towards the goal, and robustness over a single horizon.

Alg.~\ref{alg} sketches the receding-horizon procedure for planning pushing maneuvers. Starting from the initial robot configuration $\bm{q}^r_0$ and velocity $\dot{\bm{q}}^r_0$, a given initial belief $b_0$ and the number of time steps for a receding horizon $H$, BS-VP-STO is used to generate a robust pushing trajectory $\bm{q}^{r*}_{0:K}$. In order to update the belief for the subsequent receding-horizon, we perform a stochastic rollout of the robust pushing trajectory, i.e. sampling from the transition probability in \eqref{eq:transition}. Note that the update of the belief can be extended by taking observations into account. For this, line 9 in Alg.~\ref{alg} may be replaced by a Bayesian state estimation update, e.g. a particle filter (\cite{arulampalam2002tutorial}), thus turning the offline planning algorithm into an online re-planning approach. When planning offline, the optimized pushing trajectories of the receding-horizons are sequenced to form a single continuous trajectory over a long horizon. This process may be repeated until a task-specific termination criterion is satisfied, e.g. the mean object configuration is within bounds of the target.\newline

\begin{figure}[t]
    \centering
    \includegraphics[width=\linewidth]{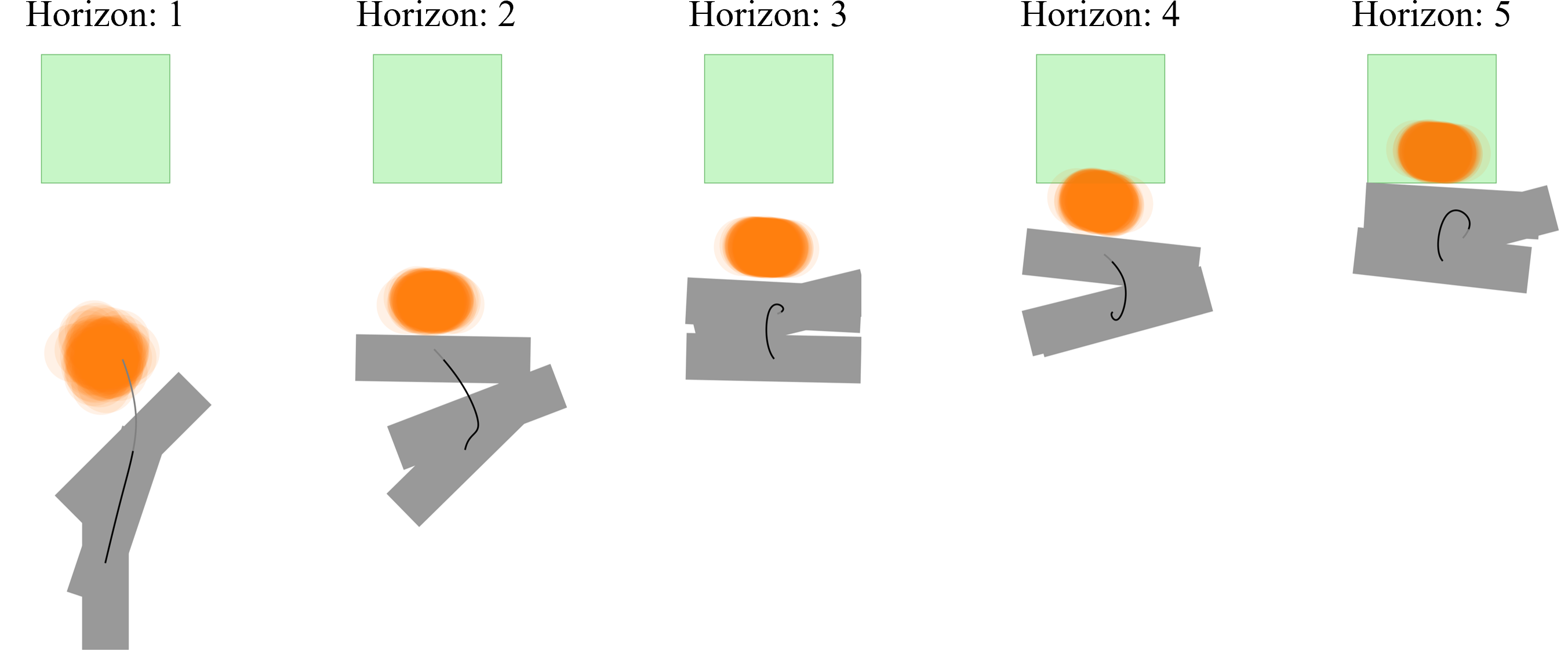}
    \caption{Receding-horizon BS-VP-STO optimizing the trajectory of a rectangular robot to push a circular object into a goal region subject to an uncertain initial object location and uncertain contact dynamics. The sub-figures show the robot trajectory optimized over the corresponding receding-horizon together with a stochastic rollout of the object dynamics. In each receding-horizon, the algorithm optimizes for a tradeoff between pushing progress and robustness. The resulting pushing maneuver is more robust than the solution found when optimizing over a single horizon as in Fig.~\ref{fig:exp_sim}.}
    \label{fig:exp_sim_mh}
\end{figure}


\noindent
\textbf{Example 3. (Receding-horizon Robust 2D Object-Pushing).} In this example, we solve the same problem as in \textbf{Example 2}, while iteratively optimizing over multiple receding horizons instead of running BS-VP-STO only once over the full horizon. For this, we only adapt the task-specific cost to reflect the pushing progress towards the goal. We measure this progress by computing the distance $d_k = ||\mathrm{E}[\bm{q}^o_k] - \bm{q}^o_{\mathrm{des}}||_2$ between the target area center $\bm{q}^o_{\mathrm{des}}$ and the mean object position at that time step. We compare the distance at the end of the receding horizon with the distance at the beginning of it in order to make the cost invariant to absolute distance to the target. Progress is thus defined as $d_0 - d_K$. Consequently, the task-cost $c_{\mathrm{task}}$ is defined as follows
\begin{equation}
    c_{\mathrm{task}}(\bm{u}_{0:K-1}) = e^{-\lambda \left( d_0 - d_K \right)}.
\end{equation}
Fig.~\ref{fig:exp_sim_mh} illustrates each solution of the optimization over multiple shorter horizons. It can be seen that a single-horizon push moves the object belief towards the target area with high probability. The overall pushing maneuver, obtained by sequencing the robot trajectories of the individual horizons, has no constraints on the number of parameters, i.e. the number of via-points, as the number of receding horizon operations is not fixed but rather tied to a goal check. It is therefore expected to be more expressive than a single-horizon solution and thus more robust.


\section{Experiments: Robust Pushing}\label{sec:experiment}

This section presents robot experiments validating the theory and algorithmic approach developed in this article. We use objects that the robot has never interacted with before, with the geometry of the objects being the only information available.
In all experiments, we compare the performance of the proposed approach against a baseline that only uses the nominal model of the contact dynamics without considering uncertainties.
For this, the planner assumes that the initial object position is known and it uses the nominal object dynamics from Sec.~\ref{sec:model} as a deterministic dynamics model. Consequently, we run the baseline without the cost and constraints on the variance gain.

\subsection{Implementation}

\begin{figure}[t]
    \centering
    \includegraphics[width=.75\linewidth]{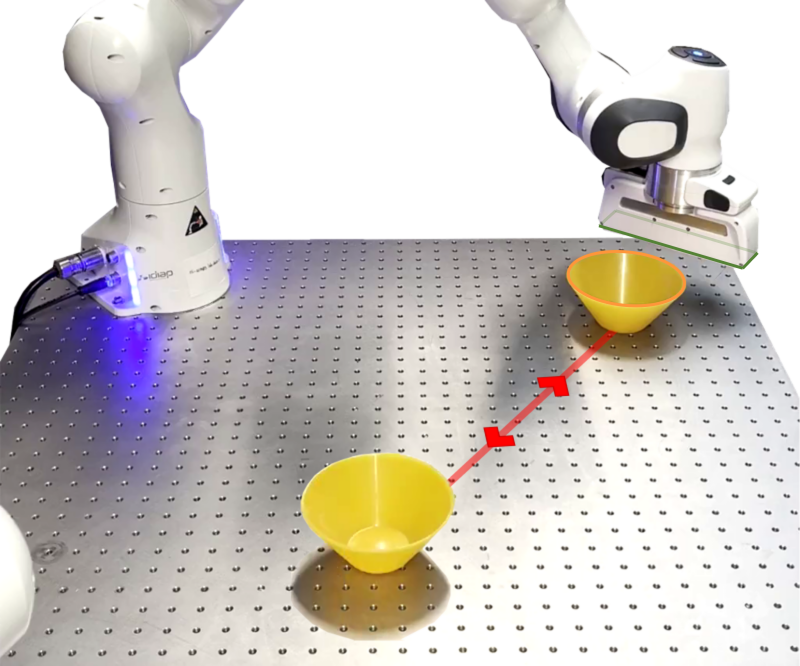}
    \caption{Open-loop single-hand pushing experiment: The task for the robot is to use the rectangular geometry of its hand (highlighted in green) to push the object with a circular geometry (highlighted in orange) into a target position without sensory feedback. The experiment consists of repeating the same pushing plan open-loop until the object diverged off the path such that the robot does not make contact anymore.}
    \label{fig:hand_pushing}
\end{figure}

We implement the contact dynamics as in \eqref{eq:object_dynamics} for the special case of circular and rectangular shapes in two dimensions. In order to approximate the initial belief via particles, we found that $N_p = 20$ particles are sufficient to generate robust plans. For computing the CMA-ES updates to the Gaussian distribution over candidate trajectories, we use the Python package provided by the authors of~\cite{Hansen2016}. 
We provide additional details on the cost design in the appendix. The planning algorithm was executed on a laptop with an Intel Core i9-14900HX CPU and 32GB of RAM. On average, generating a robust pushing plan for the full target path takes around seven seconds wall-clock time.

\subsection{Open-Loop Single-Hand Pushing}

The first experiment takes \textbf{Example 3} from simulation into the real world. As an end-effector, the robot uses the rectangular-shaped hand of the Franka robot to push the target object, without considering the fingers for contacts.
Fig.~\ref{fig:hand_pushing} illustrates the experimental setup showing the initial robot configuration, the initial object position and the target object position. We compare trajectories of 2D positions and yaw angles of the hand generated by our approach (receding-horizon BS-VP-STO) and a baseline approach. The baseline optimizes for efficient trajectories assuming that the nominal contact dynamics accurately predict the object trajectory. To evaluate the robustness of generated plans, we execute the plan to let the robot push the object into the target position and use the same plan for pushing the object back to the initial position. Subsequently, the same plan is executed repeatedly open-loop with the object located where the previous execution ended. We expect that uncertainties in the contact dynamics will lead to deviations from the nominal model, resulting in the accumulation of control errors. 
Hence, we measure robustness by counting the number of successive runs until the robot loses the object. For a video showing qualitative results, see Extension 1. In the following, we report quantitative results.

\subsubsection{Deterministic Baseline}

As the baseline does not account for uncertainties in the contact dynamics, the resulting optimal trajectory consists of pushing in a straight line towards the target position while keeping the long side of the hand perpendicular to the pushing direction. We execute 10 experiments with the optimal baseline plan, resulting in an average of $6.8$ (min. $5$, max. $9$) successive runs until the robot loses the object. This is the result of deviations from the nominal model due to real-world perturbations such as imperfect friction surfaces and mass distributions, leading to an accumulation of errors in the object positions when pushing a circular object with a flat surface.

\subsubsection{Receding-horizon BS-VP-STO}

The proposed algorithm uses the same nominal model that was used in the baseline, while modeling additional uncertainty. The resulting optimal trajectory is executed in $10$ open-loop experiments without additional perturbations to compare the result with the baseline. All experiments have been stopped after $40$ successive runs as the system did not show any sign of accumulating errors. This indicates that the optimized trajectory actively controls the uncertainty in the object position by keeping track of the open-loop propagated belief.

\subsection{Open-Loop Bi-manual Pushing}
\label{sec:exp_2}

\begin{figure*}[t]
    \centering
    \includegraphics[width=\linewidth]{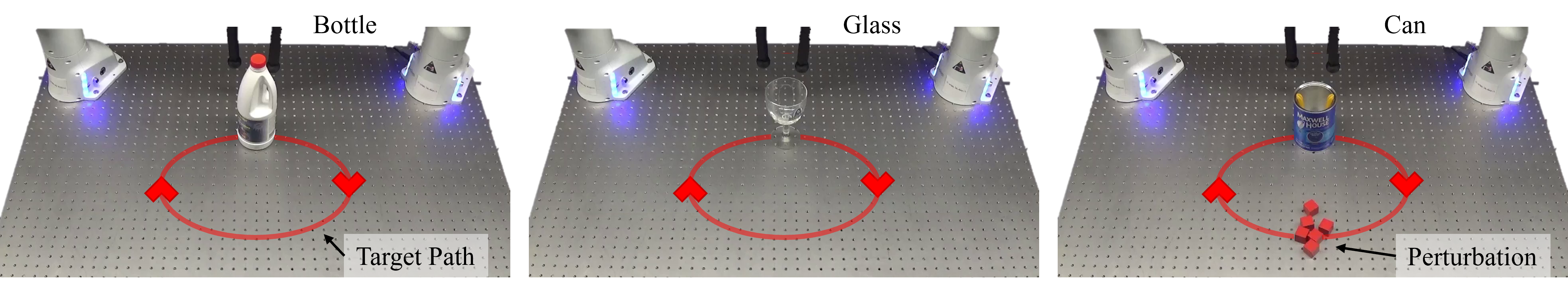}
    \caption{Open-loop bi-manual pushing experiment: The two manipulators are considered as one bi-manual robot with two end-effectors, each equipped with a ball-shaped end-effector. The object (bottle, glass or can) is placed in front of the robot and the goal is to push the object along the target path (red circle). All objects were chosen due to their circular shape for the sake of a simple implementation of the quasi-static contact dynamics. The initial belief over the object position is Gaussian distributed with a mean equal to the initial object position and a covariance that reflects the uncertainty in the object detection. The contact dynamics are modeled probabilistically to reflect uncertainty. For the experiments with the can, we add additional perturbations by placing wooden cubes (red cubes) on the target path and by changing the center of mass to be off-center by placing a heavy tool in the can (yellow content of the can).}
    \label{fig:setup}
\end{figure*}

We choose to validate the proposed algorithm with a bi-manual pushing task, consisting of two Franka robot arms that are equipped with ball-shaped end-effectors. Note that we treat the two robot arms as one bi-manual robot. We plan 2D trajectories for the ball-shaped end-effectors in a plane parallel to the table. The dynamics are modeled in this 2D plane, where the two robots are abstracted as two independent circles. The objects are abstracted as circles as well. Fig.~\ref{fig:setup} shows the experimental setup. Initially, the object (bottle, glass or can) is placed in front the bi-manual robot and the goal for the robot is to push the object along a circular target path. The belief over object positions is initialized with a Gaussian distribution. We furthermore modeled the noise in the contact dynamics with a Gaussian distribution and we tuned the covariance to capture the stochasticity in the contact dynamics. To prevent collisions between the two end-effectors, we include a collision constraint enforcing that the two end-effectors do not touch. We also add a constraint that prevents the robot from crossing its arms. 
Note that the time, location and number of contacts is subject to planning, i.e. we do not impose any heuristic that forces the robot to use both end-effectors for pushing. Instead, the proposed cost and constraint on the variance gain drives the optimization algorithm to find stabilizing contact configurations and sequences, such as using both end-effectors for pushing.


\subsubsection{Qualitative Planning Results.}

\begin{figure}[t]
    \centering
    \includegraphics[width=\linewidth]{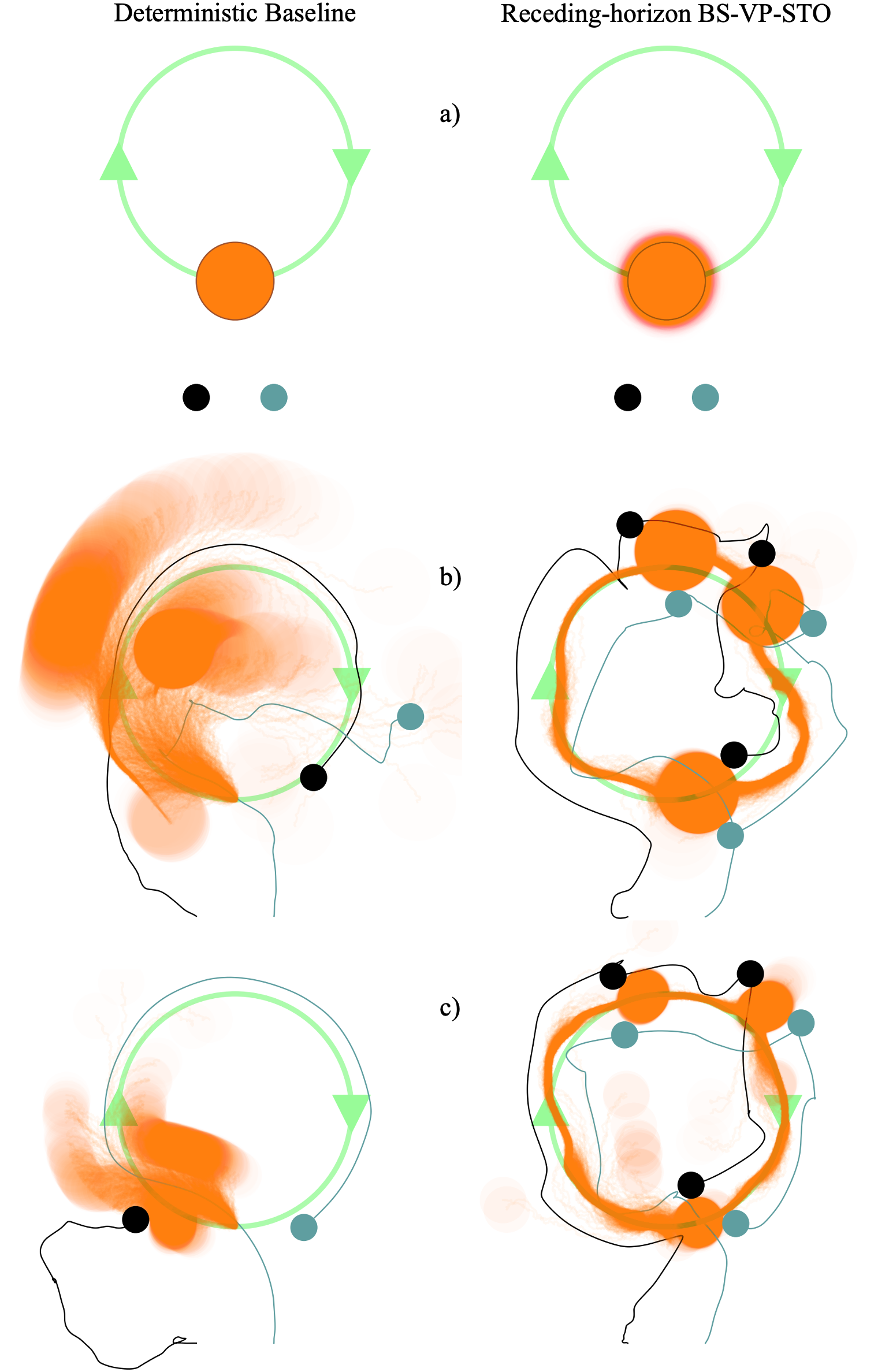}
    \caption{Qualitative comparison between the proposed receding-horizon BS-VP-STO algorithm and the deterministic baseline. We perform 1000 stochastic rollouts of each generated plan using the proposed stochastic object dynamics. The left and right end-effectors are visualized with black and blue circles, respectively. The target path is depicted in light-green. a) For testing the baseline plan, the object position is initialized with the expected position (orange circle). For the proposed approach, we initialize the object position according to the modeled uncertainty. b) Plans generated for an object with 5 cm radius, i.e. the bottle and the can. c) Plans generated for an object with 3 cm radius, i.e. the glass.}
    \label{fig:exp_plans}
\end{figure}

We illustrate plans generated with the deterministic baseline and with the proposed receding-horizon BS-VP-STO algorithm in Fig.~\ref{fig:exp_plans}. Each plan is presented with 1000 stochastic rollouts of the object dynamics used for optimization. For an evaluation of the real-world performance of the plans, please refer to Sec.~\ref{sec:real-pushing}.

\textbf{Deterministic Baseline.}
In all plans generated with the deterministic baseline, the robot uses only one end-effector at a time for pushing the object. This is not surprising as the pushing progress, i.e. the mean control problem, is equally optimized when using one or two end-effectors. Thus, the baseline is not forced to discover the coordination between the two end-effectors and converges to the simpler solution, i.e. using one end-effector. When performing open-loop stochastic rollouts of the baseline plans, the object deviates from the planned trajectory after some time and thus the robot is not able to successfully push the object along the whole target path.

\textbf{Receding-horizon BS-VP-STO.}
In contrast, we observe that the robot uses both end-effectors to push the object along the target path when planning with the receding-horizon BS-VP-STO algorithm. Note that the strategy of using two end-effectors for pushing deliberately emerges from planning in belief space. The proposed algorithm discovers the use of two end-effectors by penalizing sampled robot trajectories that result in an increasing uncertainty about the object position, i.e. using one end-effector. At the beginning of the pushing maneuver, the robot performs an action that reduces uncertainty by placing its end-effectors such that they enclose the initial belief. This effectively brings the particles closer together, resulting in a decreasing variance. The robot then starts to make contact with its two end-effectors side-by-side to push the object along the target path with high probability. After pushing the object along the first half of the circular target path, the no-collision constraint between the two end-effectors forces the robot to break the contact and to find a new contact configuration with which it can continue pushing. Consequently, the robot has to move its left end-effector around the object without touching it. Afterwards, the robot makes contact again with the object and continues pushing until it reaches the initial position again. When performing open-loop stochastic rollouts of the robust plans, the robot has a high probability of being successful at pushing the object along the target path despite the perturbations and the lack of feedback.

\subsubsection{Quantitative Planning Results \& Ablation Studies.}

\begin{figure}
    \centering
    \includegraphics[width=\linewidth]{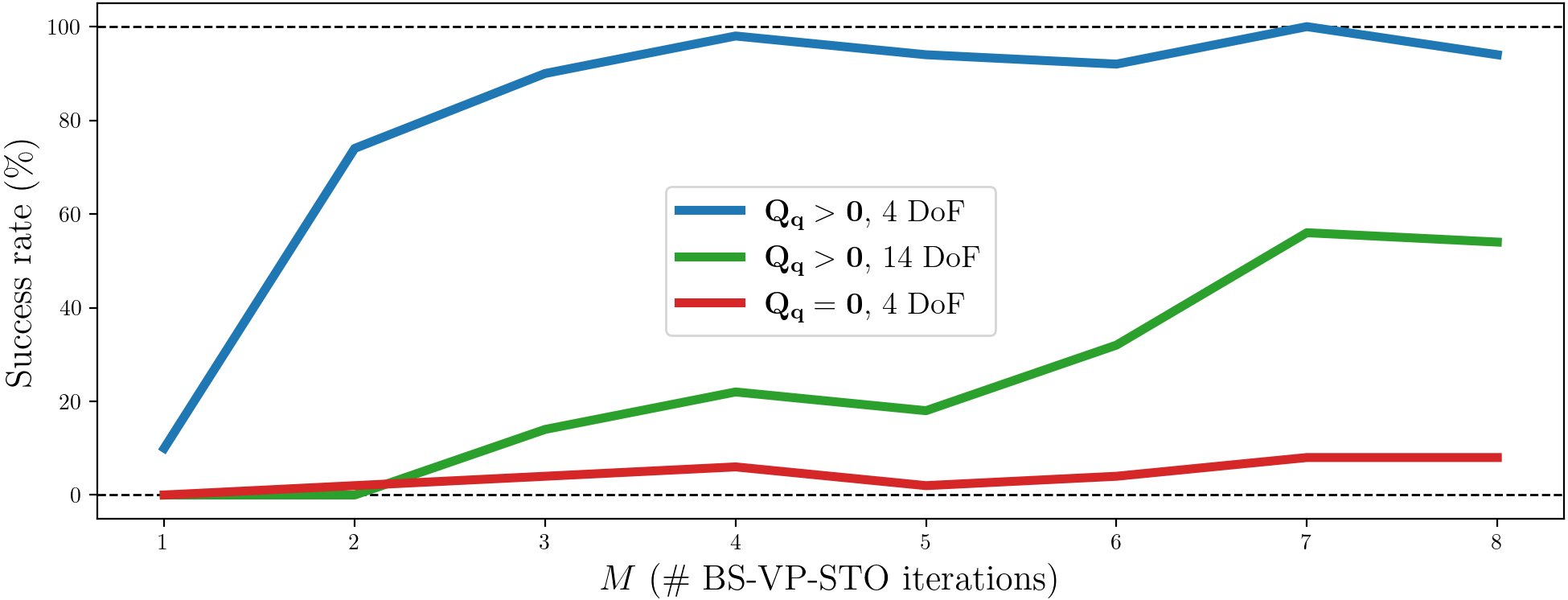}
    \caption{Success rates of receding-horizon BS-VP-STO over the number of BS-VP-STO iterations $M$. We compare the performance of the planning algorithm with the contact prior $\bm{Q}_{\bm{q}} > \bm{0}$ and without $\bm{Q}_{\bm{q}} = \bm{0}$; and we evaluate the scalability of the planning algorithm to many degrees of freedom (14 DoF). A planning run is considered successful if a valid solution is found within 500 single-horizon optimizations.}
    \label{fig:exp_stats}
\end{figure} 

In the following, we present ablation studies of the different components of receding-horizon BS-VP-STO. We evaluate \textit{i)} the dependence of the overall algorithmic performance on the number of iterations taken in each BS-VP-STO instance within the receding-horizon setting, \textit{ii)} the relevance of a contact-prior compared to an uninformed proposal distribution, and \textit{iii)} how the receding-horizon BS-VP-STO algorithm scales with the number of degrees of freedom of the robot. We summarize all findings in Fig.~\ref{fig:exp_stats} which evaluates the success rate of the receding-horizon BS-VP-STO algorithm, i.e. if the planning algorithm finds a valid solution for the robot pushing an object with 5 cm radius. A plan is considered successful if the mean of the object belief is within 1 cm tolerance to the target location and if the plan is valid with respect to the robustness constraints on the variance gain. The planning algorithm is aborted after 500 iterations of receding-horizon BS-VP-STO and the plan is considered a failure. Sub-figure b) on the right-hand side in Fig.~\ref{fig:exp_plans} illustrates a successful plan for the problem considered for the ablations. We instantiate the planning problem with a varying number $M$ of iterations for one BS-VP-STO instance within the receding-horizon scheme. For each $M$, we run receding-horizon BS-VP-STO 50 times and measure the success rate of the planning algorithm.

\begin{figure*}[th]
    \centering
    \includegraphics[width=\linewidth]{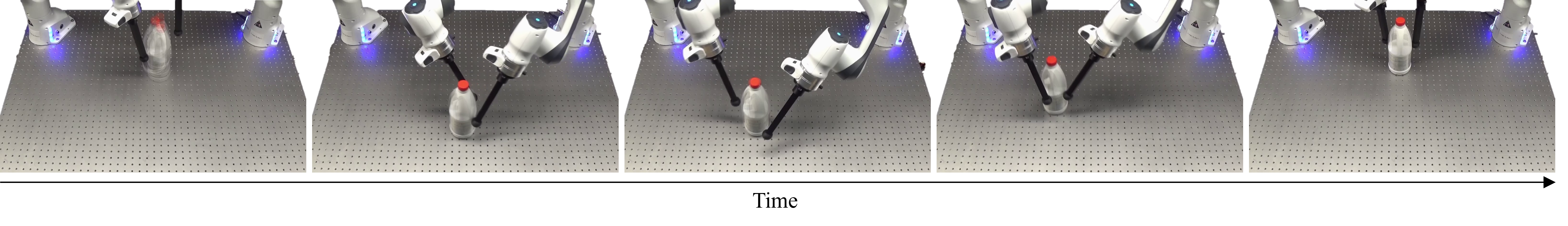}
    \caption{Snapshots of the robot behavior synthesized with the receding-horizon BS-VP-STO algorithm and executed on the real bi-manual system. Each snapshot shows an overlay of five experiments that were conducted with different initial positions of the bottle. The robot successfully pushed the bottle along the target path in all five experiments. At the beginning, i.e. in the first image, the robot encloses the initial belief with its two end-effectors such that robust pushing is possible. After moving along the first half of the circular target path, i.e. in the third image, the robot re-positions its two end-effectors to continue pushing the bottle along the target path while avoiding collisions between the two arms.}
    \label{fig:robot_snapshots}
\end{figure*}



\textbf{Impact of the Number of BS-VP-STO Iterations.}
Fig.~\ref{fig:exp_stats} illustrates the statistics of success over the number of BS-VP-STO iterations. 
We observe that the success rate increases with the number of iterations $M$ and reaches around $100 \%$ success rate with $M=4$ BS-VP-STO iterations. Note that running BS-VP-STO for one iteration poses a special case of the algorithm since no CMA-ES update is performed, resulting in only sampling an initial candidate population and picking the best performing candidate. This procedure in fact corresponds to the predictive sampling algorithm introduced in \cite{howell2022predictive}.

The planning time increases linearly with the number of iterations. For reference, performing one BS-VP-STO iteration for four degrees of freedom takes approximately 0.01 seconds wall-clock time. In this setup, we execute the pushing plan for an execution horizon $H$ that corresponds to 0.2 seconds. Thus, it would be possible to run the planning algorithm in an online receding horizon fashion with up to $M=20$ BS-VP-STO iterations per receding-horizon.

\textbf{Impact of the Contact Prior.}
To evaluate the impact of the contact prior, we set the contact precision matrix in BS-VP-STO to $\bm{Q}_{\bm{q}} = \bm{0}$ (cf. \eqref{eq:contact_prior_via}). This corresponds to uninformed trajectory samples as depicted in the left sub-figure of Fig.~\ref{fig:candidate-sampling}. In Fig.~\ref{fig:exp_stats} we observe that the success rate of the algorithm without the contact prior is significantly lower than the success rate of the algorithm with the contact prior. This indicates that the contact prior is improving the efficiency of the planning algorithm to find robust manipulation actions in few iterations. Computing the contact prior requires a matrix inversion to compute $\bm{Q}_{\bm{q}}$ with dimensionality equivalent to the number of degrees of freedom. Since this operation has to be performed only once for a single-horizon, the computational overhead of the contact prior is negligible.

\textbf{Scalability to Many Degrees of Freedom.}
Last, we evaluate the scalability of the proposed planning algorithm to many degrees of freedom by moving from planning in the 2D plane to planning in the joint space of the bi-manual robot, which has 14 degrees of freedom (seven degrees of freedom per robot arm).
Coordinating all degrees of freedom adds complexity to the planning problem, while at the same time increasing the dimensionality of the search space. In Fig.~\ref{fig:exp_stats} we observe that the success rate for a given number of iterations $M$ drops when increasing the complexity of the problem, requiring more iterations to discover the required coordination between the 14 joints. While the contact prior in joint space (cf. Sec.~\ref{sec:contact_prior_joint}) imposes coordination between the seven joints of the individual arms, BS-VP-STO is still required to find joint trajectories such that the two end-effectors touch the object at the right place at the right time. The planning time per iteration is not significantly increased compared to planning for four degrees of freedom, since only an additional forward kinematics computation for the belief rollout is required. However, note that the contact dynamics are still modeled using the ball-shaped abstraction of the end-effector. Modeling the whole kinematic chain of the robot arms as contact geometries adds additional computational complexity to the planner.

\subsubsection{Real-world Pushing Results}\label{sec:real-pushing}

Fig.~\ref{fig:robot_snapshots} shows snapshots of the robot behavior planned with the receding-horizon BS-VP-STO algorithm. In addition, a full video of the experiment can be found in Extension 2.
We executed the planned robot trajectories for each of the three objects (bottle, glass, can), where we conducted five experiments for each object by placing the object at different initial positions. Out of 15 executed plans, the robot successfully pushed the object along the target path in 14 experiments. The robot failed to push the glass in one experiment, where the object was too far away from the mean initial position such the robot did not enclose the object during the initial uncertainty-reducing action.


We furthermore evaluated the deterministic baseline planner with the same three objects, while placing the objects only at the expected location. In all three experiments, the actual object position deviated from the planned object position after a few pushes, resulting in the robot loosing contact with the object and thus failing to push the object along the target path. We show a video example of the motion generated by the baseline in Extension 2.

\subsection{Closed-Loop Bi-manual Pushing}

\begin{figure}[t!]
    \centering
    \begin{subfigure}[t]{0.5\linewidth}
        \centering
        \includegraphics[width=\linewidth]{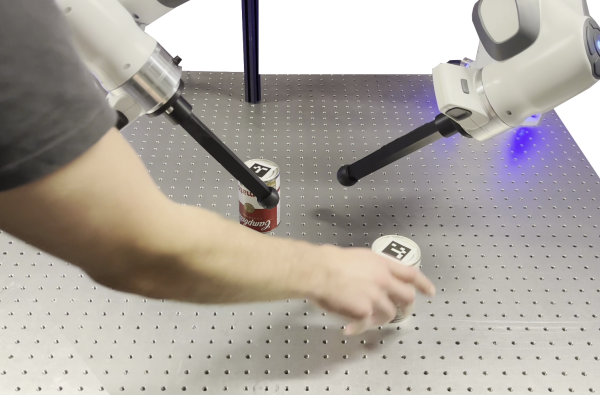}
    \end{subfigure}%
    ~ 
    \begin{subfigure}[t]{0.5\linewidth}
        \centering
        \includegraphics[width=\linewidth]{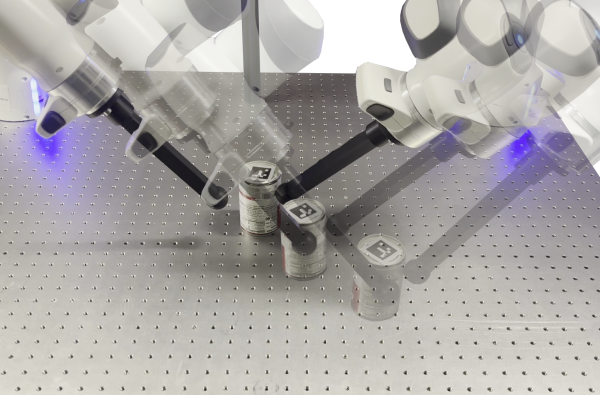}
    \end{subfigure}
    \caption{Closed-loop bi-manual pushing experiment: The left image shows the external perturbations applied by manually moving the object. A camera continuously provides noisy observations of the object position to the closed-loop controller. The right image shows how the proposed approach is able to generate robust plans in real-time, enabling uncertainty-aware model-predictive control loops.}
    \label{fig:exp_3}
\end{figure}

Last, we show that receding-horizon BS-VP-STO can be used in a closed control loop, gaining additional robustness against out-of-distribution disturbances. We use the same bi-manual robot setup as in Sec.~\ref{sec:exp_2}. However, instead of pushing an object along a target path, the task is to push the object to the center of the table while its position is perturbed by a human. In addition, noisy observations of the object position are provided by a camera for closing the loop. We deploy a particle filter for continuously updating the belief based on both the stochastic rollout and noisy observations of the object as described in Sec.~\ref{sec:multi_bsvpsto}. We qualitatively compare the resulting behavior of the robot when being controlled with receding-horizon BS-VP-STO against the behavior when using the deterministic nominal model in a model-predictive controller. Both controllers run at a control rate of $5$ Hz with $M=6$ BS-VP-STO iterations per control step. Fig.~\ref{fig:exp_3} illustrates the external perturbations applied to the object and the resulting robot behavior when controlled with our proposed robust approach. A video of the resulting behavior of both control approaches can be found in Extension 3.

We observe that the robot robustly pushes the object back to the center of the table using both end-effectors when controlled with the proposed approach. The additional state estimation enables the robot to also react to out-of-distribution perturbations, re-generating robust plans given noisy measurement updates. For the deterministic baseline, we observe that the continuous feedback enables the robot to maintain contact with the object and to generate consistent pushes. However, we observe that the robot only uses one end-effector for pushing as also observed for the open-loop experiment in Sec.~\ref{sec:exp_2}. In contrast to our approach, this leads to larger control errors during pushing that need to be corrected, resulting in the deterministic baseline taking more time to accurately bring the object back to the center of the table.





\section{Discussion}


In this article, we investigated the problem of planning robust manipulation actions subject to stochastic contact dynamics. The quasi-static model used to predict the contact dynamics is a simplification of the real-world contact dynamics tailored to the particular problem of slow pushing. In particular, reducing the dynamics from joint robot-object dynamics to solely object dynamics enables efficient reasoning over belief dynamics. However, the provided model excludes other categories of manipulation tasks involving effects such as grasping objects. We leave it to future work to investigate how other manipulation dynamics can be reduced to object-only dynamics.


Furthermore, we have shown that informed prior distributions for sampling candidate actions is beneficial, if not necessary, for sampling-based optimization for contact-rich manipulation. We used the product of Gaussian priors to bias the sampling towards smooth and contact-making trajectories. As the evolutionary optimization algorithm CMA-ES is based on sampling from and iterating on Gaussian distributions, we incorporated our Gaussian-distributed prior to initialize CMA-ES.
Yet, we see great potential in the use of non-Gaussian priors
that are further optimized with a stochastic optimization algorithm such as BS-VP-STO. Especially when the system has many degrees of freedom such as for two robot arms or articulated robotic hands, sampling from a proposal distribution that captures possibly non-Gaussian correlations between the degrees of freedom, e.g. correlations between fingers, is expected to be a key to scalable, real-time control through contacts.

Last, we show that a manipulation task such as pushing, which is typically approached with high-bandwidth closed-loop control, can also be stabilized by planning appropriate open-loop actions that deliberately optimize for robustness. However, if the robot is not able to anticipate perturbations or the statistical properties of the perturbations, feedback is required to be able to stabilize the manipulation. We thus show that the developed approach can be used in a closed-loop controller by integrating the open-loop planning algorithm in a receding-horizon control scheme and by feeding back observations of the object configuration. We furthermore show that a model-predictive controller based on receding-horizon BS-VP-STO even outperforms a controller based on a deterministic model for pushing tasks.

\section*{Declaration of conflicting interests}

The author(s) declared no potential conflicts of interest with respect to the research, authorship, and/or publication of this article.

\section*{Funding}

The author(s) disclosed receipt of the following financial support for the research, authorship, and/or publication of this article: JJ was supported by the Swiss National Science Foundation (SNSF) through the CODIMAN project. LB was supported by an Amazon Web Services Lighthouse scholarship. NH received EPSRC funding via the “From Sensing
to Collaboration” programme grant [EP/V000748/1].


\bibliographystyle{SageH}
\bibliography{references}

\section*{Appendix}

\subsection*{Separation of the Expected Cost}

Suppose that the state $\bm{x} \in \mathbb{R}^{n_x}$ that is to be controlled is a random variable that is distributed with $\bm{x} \sim p$. Furthermore suppose that the task is described by a quadratic cost of the state and a desired state $\bm{x}_{\mathrm{des}}$, i.e.
\begin{equation}
    J_{\mathrm{det}}(\bm{x}) = \left(\bm{x} - \bm{x}_{\mathrm{des}}\right)^\trsp \left(\bm{x} - \bm{x}_{\mathrm{des}}\right).
\end{equation}
The corresponding stochastic optimal control objective is given by the expectation of the quadratic cost of the state:
\begin{align}
\begin{split}
    J_{\mathrm{sto}} &= \mathrm{E}_{p} \left[ J_{\mathrm{det}}(\bm{x}) \right]\\ &= \mathrm{E}_{p} \left[ \left(\bm{x} - \bm{x}_{\mathrm{des}}\right)^\trsp \left(\bm{x} - \bm{x}_{\mathrm{des}}\right)\right]\\
    &= \mathrm{E}_{p} \left[\bm{x}^\trsp \bm{x}\right] - 2 \mathrm{E}_{p} \left[\bm{x}\right]^\trsp \bm{x}_{\mathrm{des}} + \bm{x}_{\mathrm{des}}^\trsp \bm{x}_{\mathrm{des}}.
\end{split}
\end{align}
In the following, we denote the mean of the state with
\begin{equation}
    \bar{\bm{x}} = \mathrm{E}_{p} \left[\bm{x}\right].
\end{equation}
Furthermore, the variance of the state is defined as the expectation of the squared deviations from the mean:
\begin{align}
\begin{split}
    \mathrm{V}_p\left[\bm{x}\right] &= \mathrm{E}_{p} \left[(\bm{x} - \bar{\bm{x}})^\trsp (\bm{x} - \bar{\bm{x}})\right],\\
    &= \mathrm{E}_{p} \left[\bm{x}^\trsp \bm{x} + \bar{\bm{x}}^\trsp \bar{\bm{x}} - 2 \bar{\bm{x}}^\trsp \bm{x} \right],\\
    &= \mathrm{E}_{p} \left[\bm{x}^\trsp \bm{x}\right] + \bar{\bm{x}}^\trsp \bar{\bm{x}} - 2 \bar{\bm{x}}^\trsp \mathrm{E}_{p} \left[\bm{x}\right],\\
    &= \mathrm{E}_{p} \left[\bm{x}^\trsp \bm{x}\right] - \bar{\bm{x}}^\trsp \bar{\bm{x}}.
\end{split}
\end{align}
As a result, the stochastic optimal control objective can be rewritten in terms of the deterministic quadratic cost of the mean state and the variance of the state with respect to its probability distribution $p$:
\begin{align}
\begin{split}
    J_{\mathrm{sto}} &= \bar{\bm{x}}^\trsp \bar{\bm{x}} - 2 \bar{\bm{x}}^\trsp \bm{x}_{\mathrm{des}} + \bm{x}_{\mathrm{des}}^\trsp \bm{x}_{\mathrm{des}} + \mathrm{V}_p\left[\bm{x}\right]\\
    &= \left(\bar{\bm{x}} - \bm{x}_{\mathrm{des}}\right)^\trsp \left(\bar{\bm{x}} - \bm{x}_{\mathrm{des}}\right) + \mathrm{V}_p\left[\bm{x}\right]\\
    &= J_{\mathrm{det}}(\bar{\bm{x}}) + \mathrm{V}_p\left[\bm{x}\right].
\end{split}
\end{align}

\subsection*{Smoothness Prior}

The goal of this section is to prove that $p_\mathrm{s}(\bm{q}_\mathrm{via})$ is the normalized probability density function of the unnormalized function $e^{-J_\mathrm{s}(\bm{q}_\mathrm{via})}$. $J_\mathrm{s}$ is defined with
\begin{equation}
    J_{\mathrm{s}} = \frac{1}{2} \int_0^T \ddot{\bm{q}}^{r \trsp}(t) \bm{R}_{\bm{q}} \ddot{\bm{q}}^r(t) dt.
\end{equation}
The acceleration is an affine function of the trajectory parameter, i.e.
\begin{equation}
    \ddot{\bm{q}}^r(t) = \ddot{\bm{\Phi}}_{\mathrm{via}}(t) \bm{\theta} {+} \ddot{\bm{\phi}}_{0}(t, \bm{q}_0^r, \dot{\bm{q}}_{0}^r).
\end{equation}
Now, suppose that the basis offset is computed as follows:
\begin{equation}
    \ddot{\bm{\phi}}_{0}(t, \bm{q}_0^r, \dot{\bm{q}}_{0}^r) = \ddot{\bm{\Phi}}_{\bm{q}_0}(t) \bm{q}_{0}^r + \ddot{\bm{\Phi}}_{\dot{\bm{q}}_0}(t) \dot{\bm{q}}_{0}^r.
\end{equation}
We exploit this affine dependency on the initial position and velocity by rewriting the acceleration as a linear function in a combined trajectory parameter $\bm{\xi}$, i.e.
\begin{equation}
    \ddot{\bm{q}}^r(t) = \ddot{\bm{\Phi}}(t) \bm{\xi},
\end{equation}
where the new basis function matrix and the new trajectory parameter are given by
\begin{equation}
    \ddot{\bm{\Phi}}(t) = \begin{pmatrix}
        \ddot{\bm{\Phi}}_{\mathrm{via}}(t) & \! \ddot{\bm{\Phi}}_{\bm{q}_0}(t) & \! \ddot{\bm{\Phi}}_{\dot{\bm{q}}_0}(t)
    \end{pmatrix}, \quad 
    \bm{\xi} = \begin{pmatrix}
        \bm{\theta} \\
        \bm{q}_0^r \\
        \dot{\bm{q}}_{0}^r
    \end{pmatrix}.
\end{equation}
As a result, the smoothness objective is equivalent to
\begin{align}
\label{eq:smooth_app}
\begin{split}
    J_\mathrm{s} &= \frac{1}{2} \int_0^T \bm{\xi}^\trsp \ddot{\bm{\Phi}}(t)^\trsp \bm{R}_{\bm{q}} \ddot{\bm{\Phi}}(t) \bm{\xi} \, dt,\\
    &= \frac{1}{2} \bm{\xi} \int_0^T \ddot{\bm{\Phi}}(t)^\trsp \bm{R}_{\bm{q}} \ddot{\bm{\Phi}}(t) \, dt \, \bm{\xi},\\
    &= \frac{1}{2} \bm{\xi} \bm{R}_{\bm{\xi}} \bm{\xi}.
\end{split}
\end{align}
We obtain a smoothness metric
\begin{equation}
    \bm{R}_{\bm{\xi}} = \begin{pmatrix}
        \bm{R}_{\bm{\theta}} & \bm{R}_{\bm{\theta}|\bm{q}_0,\dot{\bm{q}}_0} \\
        \bm{R}_{\bm{\theta}|\bm{q}_0,\dot{\bm{q}}_0}^\trsp & \bm{R}_{\bm{q}_0,\dot{\bm{q}}_0}
    \end{pmatrix},
\end{equation}
that is the result of the integral in the second equation in \eqref{eq:smooth_app}. With this result, the smoothness objective is in fact a quadratic function in $\bm{\xi}$ and can thus be expressed as a zero-mean Gaussian distribution with
\begin{equation}
    p_\mathrm{s}(\bm{\xi}) = p_\mathrm{s}(\bm{\theta}, \bm{q}_0^r, \dot{\bm{q}}_{0}^r) = \mathcal{N}\left(\bm{0}, \bm{R}_{\bm{\xi}}^{-1}\right).
\end{equation}
This Gaussian distribution is a joint distribution over the original trajectory parameter $\bm{\theta}$ and the initial position and velocity. At the time of constructing the smoothness prior, the initial position and velocity are given. Consequently, we obtain the smoothness prior for $\bm{\theta}$ by conditioning on the initial conditions, i.e.
\begin{align}
\begin{split}
    p_\mathrm{s}(\bm{\theta} | \bm{q}_0^r, \dot{\bm{q}}_{0}^r) = \mathcal{N}\left(\bar{\bm{\theta}}_\mathrm{s}, \bm{R}_{\bm{\theta}}^{-1}\right),\\
    \bar{\bm{\theta}}_\mathrm{s} = \bm{R}_{\bm{\theta}}^{-1} \bm{R}_{\bm{\theta}|\bm{q}_0,\dot{\bm{q}}_0} \begin{pmatrix} \bm{q}_0 \\ \dot{\bm{q}}_0 \end{pmatrix}.
\end{split}
\end{align}

\subsection*{Experiment: Implementation Details}\label{sec:app_exp}

In the following, we provide the details of the implementations for running the experiments as described in Sec.~\ref{sec:experiment}.





\subsubsection*{Path Tracking Cost.}

In a single-horizon optimization problem, we define the task-specific cost based on the object belief only. Given a predicted trajectory of particles, we denote the initial belief with $b_0$ and the belief at the end of the horizon with $b_T$. For both beliefs, we compute the expected object states, i.e. $\bar{\bm{q}}^o_0 = \mathrm{E}_{b_0}[\bm{q}^o]$ and $\bar{\bm{q}}^o_T = \mathrm{E}_{b_T}[\bm{q}^o]$, for consequently computing the cost with respect to the candidate solution. The target path is defined as a circle with
\begin{equation}
    \bm{q}^o_{\mathrm{path}}(s) = r_{\mathrm{path}} \begin{pmatrix}
        \cos (2\pi s) \\ \sin (2\pi s)
    \end{pmatrix}, \, s \in [0,1].
\end{equation}
The radius of the circle is $r_{\mathrm{path}} = 0.15$.
Next, we find the point on the target path that is closest to the expected object states with
\begin{equation}
    s_k = \argmin_s ||\bar{\bm{q}}^o_k - \bm{q}^o_{\mathrm{path}}(s)||_2, \, \mathrm{s.t.} \, s \in [0,1],
\end{equation}
for $k = \{0,T\}$. Consequently, we compute the path tracking cost with
\begin{equation}
    c_{\mathrm{path}} = e^{w_\mathrm{progress} (s_0 - s_T)} + w_\mathrm{error} ||\bar{\bm{q}}^o_T - \bm{q}^o_{\mathrm{path}}(s_T)||_2^2.
\end{equation}
The first cost term measures the tracking progress by comparing the progress variable $s$ at the end of the horizon with the beginning of the horizon. We scale the first cost term with $w_\mathrm{progress} = 100$. The second cost term penalizes deviations from the path, where the squared error term is weighted with $w_\mathrm{error} = 2000$.

\end{document}